\documentclass[10pt,twocolumn,letterpaper]{article}

\usepackage{cvpr}              
\usepackage[accsupp]{axessibility}

\usepackage[utf8]{inputenc} 
\usepackage[T1]{fontenc}    
\usepackage{url}            
\usepackage{booktabs}       
\usepackage{amsfonts}       
\usepackage{nicefrac}       
\usepackage{microtype}      
\usepackage{xcolor}         
\usepackage{graphicx}
\usepackage{amsmath}
\usepackage{amssymb}
\usepackage{amsthm}
\usepackage{algorithm}
\usepackage{algpseudocode}
\usepackage{booktabs}
\usepackage{enumitem}
\usepackage{multirow}
\usepackage{multicol}
\usepackage{pifont}
\usepackage{xspace}

\makeatletter
\DeclareRobustCommand\onedot{\futurelet\@let@token\@onedot}
\def\@onedot{\ifx\@let@token.\else.\null\fi\xspace}
\def\eg{\emph{e.g}\onedot}

\makeatother

\newcommand*\colourcheck[1]{%
  \expandafter\newcommand\csname #1check\endcsname{\textcolor{#1}{\ding{52}}}%
}
\colourcheck{green}
\newcommand*\colourxmark[1]{%
  \expandafter\newcommand\csname #1xmark\endcsname{\textcolor{#1}{\ding{55}}}%
}
\colourxmark{red}

\theoremstyle{plain}
\newtheorem{theorem}{Theorem}

\theoremstyle{definition}

\theoremstyle{remark}

\definecolor{cvprblue}{rgb}{0.21,0.49,0.74}
\usepackage[pagebackref,breaklinks,colorlinks,allcolors=cvprblue]{hyperref}

\usepackage[capitalize]{cleveref}
\crefname{section}{Sec.}{Secs.}
\Crefname{section}{Section}{Sections}
\Crefname{table}{Table}{Tables}
\crefname{table}{Tab.}{Tabs.}
\crefname{theorem}{Theorem}{Theorems}

\def\confName{CVPR}
\def\confYear{2026}

\title{Generalized and Personalized Federated Learning\\with Black-Box Foundation Models via Orthogonal Transformations}

\author{
Eun Gyung Kong\thanks{Equal contribution.}~~\thanks{Current affiliation: Mobilint, Inc.} \quad 
Je Won Yeom$^{*}$ \quad 
Yonghoon Jeon\thanks{Current affiliation: Kakao Healthcare Corp.} \quad 
Taesup Kim\thanks{Corresponding author.} \\
Seoul National University\\
}

\begin{document}
\maketitle

\begin{abstract}
Federated Learning (FL) facilitates decentralized model training while preserving data privacy. However, achieving both robust generalization and effective personalization simultaneously in heterogeneous (non-IID) environments remains a formidable challenge. Furthermore, the widespread adoption of proprietary Foundation Models (FMs) introduces a critical requirement for \textit{dual privacy}: (a) protecting sensitive client data and (b) securing the server's valuable intellectual property. This mandates strictly black-box access to the FM. To address these multifaceted challenges, we introduce \textsc{FedOT}, a novel FL framework optimized for black-box FMs. \textsc{FedOT} employs a shared global task-dependent classifier while facilitating local adaptation through client-specific orthogonal transformations applied externally to the FM embeddings. This architecture inherently guarantees that the FM's internal parameters remain inaccessible and unmodified. By enforcing orthogonality, \textsc{FedOT} effectively mitigates gradient conflicts across diverse clients, which is theoretically bounded, preserves the semantic integrity of the FM representations, and achieves robust performance under significant data heterogeneity. The synergy of global and local parameters optimally balances generalization and personalization, markedly outperforming baseline FL methods across diverse benchmarks. Extensive empirical analysis, including rigorous multi-seed validation and scalability assessments, substantiates the robustness, efficiency, and superior performance of \textsc{FedOT}.
\end{abstract}

\section{Introduction}
\label{sec:intro}

The rapid evolution of deep learning underscores the necessity of leveraging large-scale datasets for training high-performance models. However, the centralization of such data is increasingly infeasible due to stringent privacy regulations and heightened concerns regarding data sensitivity.
Federated Learning (FL) emerges as a compelling paradigm by enabling decentralized model training across distributed clients without necessitating centralized data aggregation~\cite{RN705,RN706,mcmahan2017communication}. While the primary goal of FL is often the development of a generalized model applicable to both existing and future clients, significant variations in local data distributions (non-IID settings) frequently demand personalized models. This personalization is particularly crucial in high-stakes domains such as healthcare~\cite{lu2022personalized}. Consequently, achieving an optimal equilibrium between generalization and personalization under severe data heterogeneity remains a fundamental challenge in FL research.

\begin{figure*}[t]
\centering
\includegraphics[width=0.8\textwidth]{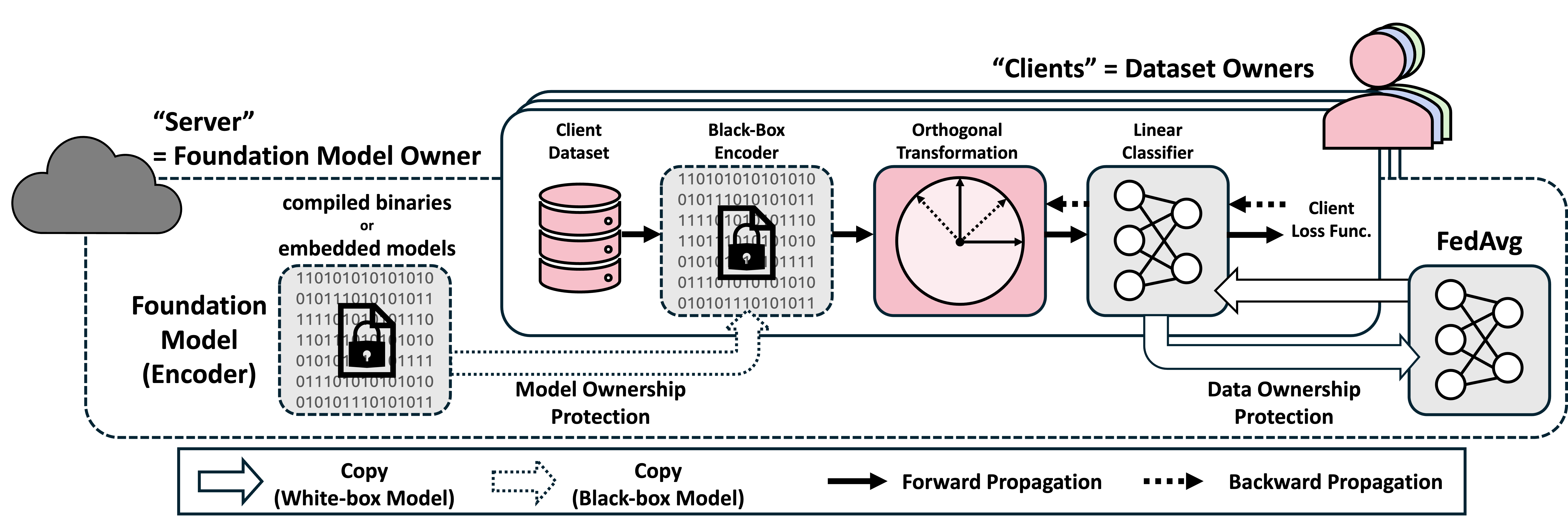}
\caption{\textbf{Overview of our proposed FL framework, \textsc{FedOT}.} We leverage a pre-trained vision encoder in a strictly black-box FL environment. The image encoder, deployed via API or as a compiled binary, operates with client-specific orthogonal transformations (local parameters) applied externally to the feature embeddings for personalization, while utilizing a globally shared classifier for generalization. This design ensures \textit{dual privacy}: protecting client data and safeguarding the server's model IP, as the FM internals are never accessed or modified.}
\label{fig:overview}
\end{figure*}

Recently, foundation models (FMs)~\cite{Bommasani2021FM}, including advanced vision-language models (VLMs) like CLIP~\cite{RN568}, have exhibited remarkable generalization capabilities, making their integration into FL architectures highly desirable.
However, deploying proprietary FMs in practical FL scenarios introduces a novel and critical challenge: \textit{dual privacy}. While FL inherently protects client data by maintaining locality, the model provider (server) also possesses strong incentives to protect the intellectual property (IP) of their high-value FMs. This imperative often dictates that clients interact with the FM in a strictly \textit{black-box} manner (\eg, via API calls or deployed binaries), without access to internal weights, architectures, or gradients~\cite{xiao2023offsite, lu2024zoopfl}.

This black-box constraint renders many conventional parameter-efficient fine-tuning (PEFT) methods unsuitable. Techniques such as LoRA~\cite{RN771} or Adapters, along with numerous established Personalized FL (PFL) approaches, typically necessitate \textit{white-box} access. They rely on inserting trainable modules deep within the network architecture or accessing internal gradients for updates~\cite{RN683, RN682}. These requirements are fundamentally incompatible with the stringent security protocols governing proprietary FM deployment.

To address this critical gap, we propose \textsc{FedOT} (Federated Learning via Orthogonal Transformations), a simple yet highly effective framework specifically engineered for the constraints of black-box FMs. In \textsc{FedOT}, clients collaboratively train and communicate only a task-dependent global classifier. Concurrently, they personalize their models by locally adapting the visual features using orthogonal transformations applied \textit{externally} to the FM embeddings. This approach guarantees both model security (server IP) and data security (client privacy) by design.

The core innovation of \textsc{FedOT} lies in the synergistic integration of these components, carefully tailored for this constrained environment. We demonstrate that orthogonal transformations are uniquely suited for this task, as they mitigate gradient conflicts across heterogeneous clients while preserving the intrinsic geometric properties of the FM's representation space. We theoretically prove that orthogonal transformations, characterized by a condition number of 1 ($\kappa = 1$), achieve the smallest possible upper bound on gradient discrepancy across clients (\cref{thm:GradDiffBound}), thereby stabilizing training and enhancing generalization. Our extensive empirical evaluation, rigorously validated across multiple random seeds and scaled up to 75 clients, demonstrates that \textsc{FedOT} consistently delivers superior performance and confirms the critical role of the orthogonality constraint.

The contributions of our framework are summarized as follows:
\begin{itemize} [leftmargin=1.5em]
    \item We introduce \textsc{FedOT}, a novel and practical framework specifically designed for Federated Learning with \textit{black-box} Foundation Models. It ensures \textit{dual privacy} by operating entirely on external embeddings without requiring any internal model access.
    \item To the best of our knowledge, this work is the first to theoretically establish and empirically validate the effectiveness of Orthogonal Transformations in minimizing gradient conflicts and stabilizing training within this specific black-box FL context.
    \item Our framework is validated through comprehensive empirical evaluations, including large-scale scalability tests up to 75 clients and rigorous multi-seed validation. \textsc{FedOT} consistently surpasses existing approaches in achieving an optimal balance between generalization and personalization across diverse heterogeneous datasets.
\end{itemize}

\section{Related Work} \label{sec:relatedwork}

\paragraph{Domain Generalization and Adaptation.} Domain Generalization (DG) and Domain Adaptation (DA) strategies aim to enhance model robustness against distribution shifts. DG focuses on training across multiple source domains to generalize to unseen target domains \cite{RN521}, often leveraging domain-invariant representations \cite{RN650,RN547} or meta-learning techniques \cite{RN536, RN654}. DA typically involves fine-tuning source-pre-trained models using limited target data \cite{RN528}. While recent advancements explore test-time adaptation \cite{Chen2023ITTA, RN656}, the majority of these solutions assume centralized data access, a limitation addressed by emerging Federated DG frameworks \cite{Zhang2023FedDG, Gong2024PLAN}.

\paragraph{Orthogonality in Fine-Tuning.} Recent studies have investigated the role of orthogonality in the efficient adaptation of pre-trained models. \textsc{Pro$^2$}~\cite{chen2023project} employs a linear projection constructed with an orthonormal basis for adaptation. Orthogonal Fine-Tuning (OFT)~\cite{RN683, RN682} provides an alternative to low-rank adapters such as LoRA~\cite{RN771} by reparameterizing pre-trained weights using orthogonal matrices. OFT has been shown to preserve semantic structure by maintaining hyper-spherical energy. While effective, these methodologies inherently require \textit{white-box} access to the model architecture. They involve modifications to internal mechanisms, such as the reparameterization of existing weight matrices, making them unsuitable for black-box scenarios.

\paragraph{Black-Box Foundation Models and Federated Learning.} Foundation models are increasingly deployed under strict black-box constraints due to intellectual property protection concerns~\cite{sun2022black,yu2023black,paranjape2024blackboxadapt}. This constraint renders many traditional PFL methods, which assume \textit{white-box} access for fine-tuning feature extractors or inserting internal modules, inapplicable. Similarly, the \textit{white-box} orthogonal methods discussed above are also incompatible. Common black-box adaptation strategies include gradient-free prompt tuning~\cite{sun2022black,yu2023black}, the utilization of external wrappers or modules~\cite{ormazabal2023comblm,zhang2022tip}, and embedding-based feature transformations~\cite{ouali2023blackboxfewshot,xiao2023offsite}. In the context of Federated Learning (FL), ZooPFL~\cite{lu2024zoopfl} introduced the challenge of black-box foundation FL, addressing personalization by adapting model \textit{inputs} via zeroth-order optimization. While innovative, this approach can be computationally intensive due to the high query complexity required for gradient estimation.~\cite{lu2024zoopfl} \textsc{FedOT} distinguishes itself by employing efficient gradient-based learning through the transformation of \textit{output features} via orthogonal transformations. This approach is fully decoupled from the FM, effectively supporting both personalization and generalization while maintaining high computational efficiency and ensuring dual privacy.

\section{Method}
\label{sec:method}
We introduce \textsc{FedOT}, a parameter-efficient framework for FL designed to preserve client data privacy and protect proprietary foundation models (dual privacy), while simultaneously enhancing both generalization and personalization capabilities.

\subsection{Problem Formulation: FL with Black-Box FMs}
We adopt the definitions of generalization and personalization in the FL context as established by FedCLIP~\cite{RN587}.
We consider an FL scenario where a set of diverse clients actively participate in the training process. This motivates the need for personalization to each client, while anticipating the deployment of the model to new, unseen clients after the learning process concludes, which motivates the need for generalization to unseen client.
Crucially, we assume a black-box vision FM that serves strictly as a fixed image encoder $\mathcal{I}(\cdot)$. Clients are restricted to accessing only the output embeddings and are prohibited from modifying the internal parameters of any FM used in our system.

\subsection{Global and Local Parameters}
\textsc{FedOT} integrates both global and local parameters within the FL architecture, leveraging the representational power of FMs without violating the black-box constraint (\cref{fig:overview}).

\paragraph{Local Parameters: Personalized Orthogonal Transformations.} We introduce client-specific local parameters in the form of \textit{orthogonal transformations} $w_\mathrm{l}^{(i)}\in\mathbb{R}^{d\times d}$ for client $i$. These parameters enable personalization by adapting representations externally to the FM. Orthogonal transformations are central to our approach. They are isometric, preserving the geometric properties (specifically, lengths and angles) of the vector space. This ensures that the semantic integrity and the underlying manifold structure of the original FM features are maintained, while allowing for necessary client-specific adaptations.

\paragraph{Global Parameters: Shared Generalized Classifier.} To foster generalization across heterogeneous domains, we employ a \textit{global classifier} $w_{\mathrm{g}} \in \mathbb{R}^{K \times d}$ (where $K$ is the number of classes). This classifier is collaboratively trained and shared among all participating clients.
It can be randomly initialized or, when leveraging VLMs, initialized from class embeddings $t=\{\mathcal{T}(p_c)\}_{c=1}^{K}$ using the VLM text encoder $\mathcal{T}$ (utilized only once during initialization).\\ \\
This dual-parameter architecture simultaneously achieves robust generalization across clients and tailored personalization for each client's unique data distribution.

\subsection{Parameter Update and Optimization}
During initialization, global parameters are distributed $w_{\mathrm{g}}^{(i)} \leftarrow w_{\mathrm{g}}$, and local parameters are initialized to the identity matrix $w_{\mathrm{l}}^{(i)} \leftarrow I$.

The original visual feature $h=\mathcal{I}(x)$ is linearly transformed into $h^{\prime} = w_{\mathrm{l}}^{(i)}h$. For client $i$, the classification probability is calculated as:
\begin{equation*}
P(y\mid x; w_\mathrm{g}, w_\mathrm{l}^{(i)}) = \mathrm{softmax}\left(\tau w_\mathrm{g} \left({h^{\prime}}/{\|h^{\prime}\|}\right)\right),
\label{eqn:prob}
\end{equation*}
where $\tau$ is a temperature hyperparameter. Each client \(i\) minimizes the standard cross-entropy loss:
\begin{equation*}
\ell^{(i)}\bigl(w_{\mathrm{g}}^{(i)},\,w_{\mathrm{l}}^{(i)}\bigr)
=
\mathbb{E}_{(x,y)\sim D^{(i)}}\bigl[-\log P\bigl(y\mid x; w_\mathrm{g}^{(i)}, w_\mathrm{l}^{(i)}\bigr)\bigr].
\end{equation*}

\paragraph{Optimization via the Cayley Transform.}
To enforce orthogonality during gradient-based optimization, we employ the differentiable \emph{Cayley transform}. This method provides a smooth parameterization of the Stiefel manifold, ensuring stability during SGD updates. We optimize an unconstrained matrix \(X^{(i)}\) and derive the orthogonal transformation \(w_{\mathrm{l}}^{(i)}\) from it. Specifically:
$w_{\mathrm{l}}^{(i)} = \bigl(I + R^{(i)}\bigr)\,\bigl(I - R^{(i)}\bigr)^{-1}$, where $R^{(i)} = \frac{1}{2}\bigl(X^{(i)} - (X^{(i)})^\top\bigr)$ is a skew-symmetric matrix derived from $X^{(i)}$. This parameterization guarantees that $w_{\mathrm{l}}^{(i)}$ remains strictly orthogonal throughout the training process.

\paragraph{Algorithm Overview.}
During local training, both $w_{\mathrm{g}}^{(i)}$ and $X^{(i)}$ are updated via SGD. $w_{\mathrm{l}}^{(i)}$ is then recomputed from the updated $X^{(i)}$ using the Cayley transform, maintaining orthogonality at each step.

Upon completion of local training, the server aggregates only the global parameters: $w_{\mathrm{g}} \leftarrow \frac{1}{N}\,\sum_{i=1}^{N} w_{\mathrm{g}}^{(i)}$. Refer to Algorithm 1
in the Appendix for the complete pseudo-code.

\subsection{Protecting Proprietary Models and Data}
\label{method:protect}
\textsc{FedOT} ensures \textit{dual privacy} by architectural design.
\textbf{Client data privacy} is protected as raw data remains strictly local, and the personalized local transformations ($w_\mathrm{l}^{(i)}$) are never transmitted to the server.
\textbf{Server model IP} is secured because the adaptation mechanism operates entirely externally on the extracted image features. It requires neither access to, nor modification of, the internal parameters of the image encoder. This enables the FM to function as a strictly black-box module (\eg, compiled binary or API endpoint).

\section{Theoretical Analysis of FedOT}
\label{sec:analysis}

We present theoretical insights demonstrating how the enforcement of orthogonality on the local parameters \(w_{\mathrm{l}}^{(i)}\) effectively mitigates gradient conflicts among heterogeneous clients, thereby stabilizing the federated learning process.

\subsection{Bounding the Gradient Difference}
We compute the derivative of the cross-entropy loss \(\ell^{(i)}\) with respect to the global parameter \(w_{\mathrm{g}}^{(i)}\):
\begin{equation*}
\nabla_{w_{\mathrm{g}}^{(i)}} \ell^{(i)}
=
\mathbb{E}\Bigl[
\tau\,\bigl(r_{x}^{(i)} - y\bigr)\,
\bigl(\tfrac{h^{\prime}}{\|h^{\prime}\|}\bigr)^\top
\Bigr],
\end{equation*}
where \(r_{x}^{(i)}\) denotes the classification probability vector. We investigate how the discrepancy between these gradients across two distinct clients \(i\) and \(j\) can be formally bounded.

\begin{theorem}
\label{thm:GradDiffBound}
(Gradient Difference Bound)
Let \(i\) and \(j\) be two distinct clients. Then the \(\ell_2\)-norm of the difference between their global-parameter gradients satisfies
\begin{equation*}
\Bigl\|
\nabla_{w_{\mathrm{g}}^{(i)}} \ell^{(i)}
-
\nabla_{w_{\mathrm{g}}^{(j)}} \ell^{(j)}
\Bigr\|
\le
2\tau\,\Bigl[\kappa\bigl(w_{\mathrm{l}}^{(i)}\bigr) + \kappa\bigl(w_{\mathrm{l}}^{(j)}\bigr)\Bigr],
\end{equation*}
where \(\kappa(\cdot)\) denotes the condition number of the corresponding linear transformation. Moreover, if \(\kappa\bigl(w_{\mathrm{l}}^{(i)}\bigr) = \kappa\bigl(w_{\mathrm{l}}^{(j)}\bigr) = 1\), the bound tightens to \(4\tau\), which is the smallest achievable value for this upper bound. For a detailed proof, see 
Sec. C in the Appendix.
\end{theorem}

\begin{table*}[t]
\caption{\textbf{Comparison of generalization (G), personalization (P), and comprehensive (C) accuracy (\%).} We report the mean and standard deviation across 3 random seeds (50, 77, 98) for FEMNIST, PACS, OfficeHome, VLCS, and TerraIncognita, and the average across them. The best results are highlighted in \textbf{bold}. \textsc{FedOT} demonstrates a superior balance across all metrics.}
\label{tab:result}
\centering
\resizebox{\textwidth}{!}{
\begin{tabular}{c}
\begin{tabular}{lccccccccc}
\toprule
\multirow{2}{*}{Method}
& \multicolumn{3}{c}{FEMNIST}& \multicolumn{3}{c}{PACS}& \multicolumn{3}{c}{Office-Home} \\
\cmidrule(lr){2-4}\cmidrule(lr){5-7}\cmidrule(lr){8-10}
& G(\%) & P(\%) & C(\%)
& G(\%) & P(\%) & C(\%)
& G(\%) & P(\%) & C(\%) \\
\midrule
\midrule
CLIP (ZS)
& 48.18 $\pm$ 1.31 & -- & --
& 94.36 $\pm$ 0.48 & -- & --
& 79.15 $\pm$ 0.40 & -- & -- \\
\midrule
FedCLIP
& 70.55 $\pm$ 8.60 & 70.49 $\pm$ 9.54 & 70.51 $\pm$ 9.30
& 94.24 $\pm$ 0.71 & 95.82 $\pm$ 0.79 & 95.42 $\pm$ 0.76
& 78.93 $\pm$ 1.06 & 80.51 $\pm$ 0.26 & 80.12 $\pm$ 0.17 \\
PromptFL
& 96.29 $\pm$ 0.86 & 96.44 $\pm$ 0.70 & 96.40 $\pm$ 0.74
& 94.06 $\pm$ 0.61 & 96.96 $\pm$ 0.35 & 96.23 $\pm$ 0.37
& 80.42 $\pm$ 0.80 & 84.27 $\pm$ 0.89 & 83.31 $\pm$ 0.86 \\
CoCoOp
& 14.91 $\pm$ 3.27 & 95.38 $\pm$ 0.08 & 75.26 $\pm$ 0.82
& 82.49 $\pm$ 8.16 & 96.32 $\pm$ 0.50 & 92.86 $\pm$ 2.22
& 72.73 $\pm$ 1.46 & 81.45 $\pm$ 0.45 & 79.27 $\pm$ 0.70 \\
VPT
& 65.73 $\pm$ 6.97 & 91.13 $\pm$ 0.93 & 84.78 $\pm$ 2.39
& 94.50 $\pm$ 0.59 & 96.60 $\pm$ 0.44 & 96.07 $\pm$ 0.47
& 80.36 $\pm$ 0.74 & 82.90 $\pm$ 0.36 & 81.52 $\pm$ 0.45 \\
\midrule
FedAKT(C)
& 9.13 $\pm$ 3.53 & 55.02 $\pm$ 9.71 & 45.01 $\pm$ 6.17
& 14.07 $\pm$ 1.19 & 73.06 $\pm$ 3.68 & 60.43 $\pm$ 1.16
& 0.87 $\pm$ 0.09 & 73.49 $\pm$ 0.46 & 55.34 $\pm$ 0.37 \\
FedGH(C)
& 39.29 $\pm$ 3.34 & 39.55 $\pm$ 1.04 & 39.38 $\pm$ 0.58
& 55.23 $\pm$ 1.62 & 63.44 $\pm$ 1.59 & 61.49 $\pm$ 1.48
& 1.91 $\pm$ 0.13 & 8.70 $\pm$ 3.05 & 7.00 $\pm$ 2.27 \\
\midrule
FedLT
& 88.74 $\pm$ 0.14 & 96.61 $\pm$ 0.36 & 94.65 $\pm$ 0.30
& 94.32 $\pm$ 0.40 & 97.23 $\pm$ 0.20 & 96.50 $\pm$ 0.24
& 80.57 $\pm$ 0.31 & 87.01 $\pm$ 0.34 & 85.40 $\pm$ 0.32 \\
FedAdapter
& 96.10 $\pm$ 0.47 & 96.38 $\pm$ 0.28 & 96.31 $\pm$ 0.33
& 94.37 $\pm$ 0.45 & 97.01 $\pm$ 0.16 & 96.35 $\pm$ 0.23
& \textbf{81.33 $\pm$ 0.26} & 85.96 $\pm$ 0.51 & 84.80 $\pm$ 0.45 \\
\midrule
FedOT(All Global)
& 93.31 $\pm$ 3.04 & 93.31 $\pm$ 3.23 & 93.31 $\pm$ 3.17
& \textbf{94.55 $\pm$ 0.54} & 96.85 $\pm$ 0.39 & 96.28 $\pm$ 0.43
& 81.07 $\pm$ 0.85 & 83.80 $\pm$ 0.81 & 83.12 $\pm$ 0.81 \\
FedOT(All Local) 
& -- & 87.74 $\pm$ 0.80 & --
& -- & 97.04 $\pm$ 0.49 & --
& -- &84.70 $\pm$ 0.37 & -- \\
\midrule
\textsc{FedOT} (Ours)
& 94.23 $\pm$ 0.84 & 95.18 $\pm$ 1.10 & 94.94 $\pm$ 1.01
& 94.53 $\pm$ 0.35 & \textbf{97.34 $\pm$ 0.23} & \textbf{96.64 $\pm$ 0.25}
& 80.59 $\pm$ 0.47 & \textbf{87.17 $\pm$ 0.27} & \textbf{85.53 $\pm$ 0.32}  \\
\textsc{FedOT}(+B) (Ours)
& \textbf{97.51 $\pm$ 0.48} & \textbf{97.64 $\pm$ 0.45} & \textbf{97.61 $\pm$ 0.44}
& 94.53 $\pm$ 0.35 & \textbf{97.34 $\pm$ 0.23} & \textbf{96.64 $\pm$ 0.25}
& 80.59 $\pm$ 0.47 & \textbf{87.17 $\pm$ 0.27} & \textbf{85.53 $\pm$ 0.32} \\
\bottomrule\end{tabular}
\\ \vspace{0.01cm} \\
\begin{tabular}{lcccccc|ccc}
\toprule
\multirow{2}{*}{Method}
& \multicolumn{3}{c}{VLCS}& \multicolumn{3}{c}{TerraIncognita}& \multicolumn{3}{|c}{Average} \\
\cmidrule(lr){2-4}\cmidrule(lr){5-7}\cmidrule(lr){8-10}
& G(\%) & P(\%) & C(\%)
& G(\%) & P(\%) & C(\%)
& G(\%) & P(\%) & C(\%) \\
\midrule
\midrule
CLIP (ZS)
& 80.59 $\pm$ 0.96 & -- & --
& 19.08 $\pm$ 0.26 & -- & --
& 67.31 & -- & -- \\
\midrule
FedCLIP
& 80.45 $\pm$ 0.86 & 84.92 $\pm$ 0.87 & 83.80 $\pm$ 0.63
& 34.62 $\pm$ 1.63 & 54.05 $\pm$ 12.04 & 49.19 $\pm$ 9.11
& 71.76 & 77.16 & 75.81 \\
PromptFL
& 81.79 $\pm$ 0.15 & 87.09 $\pm$ 0.80 & 85.77 $\pm$ 0.57
& 37.51 $\pm$ 1.92 & 60.07 $\pm$ 0.60 & 54.43 $\pm$ 0.75
& 78.01 & 84.97 & 83.23 \\
CoCoOp
& 69.60 $\pm$ 7.67 & 87.49 $\pm$ 0.66 & 83.02 $\pm$ 1.46
& 26.17 $\pm$ 5.38 & 66.80 $\pm$ 1.69 & 56.65 $\pm$ 2.22
& 53.18 & 85.49 & 77.41 \\
VPT
& 80.89 $\pm$ 1.02 & 87.71 $\pm$ 1.16 & 86.00 $\pm$ 1.08
& 34.76 $\pm$ 0.45 & 58.45 $\pm$ 4.86 & 52.53 $\pm$ 3.71
& 71.25 & 83.36 & 80.18 \\
\midrule
FedAKT(C)
& 11.54 $\pm$ 1.30 & 88.39 $\pm$ 0.40 & 69.18 $\pm$ 0.44
& 12.25 $\pm$ 5.78 & 47.43 $\pm$ 3.10 & 39.44 $\pm$ 0.91
& 9.57 & 67.48 & 53.88 \\
FedGH(C)
& 47.43 $\pm$ 2.02 & 59.21 $\pm$ 1.84 & 56.27 $\pm$ 1.38
& 7.28 $\pm$ 2.80 & 10.66 $\pm$ 7.59 & 10.12 $\pm$ 5.98
& 30.23 & 36.31 & 34.85 \\
\midrule
FedLT
& 82.21 $\pm$ 0.44 & 88.97 $\pm$ 0.31 & 87.28 $\pm$ 0.34
& 29.87 $\pm$ 2.29 & 73.43 $\pm$ 0.88 & 62.54 $\pm$ 0.09
& 74.77 & 88.43 & 84.24 \\
FedAdapter
& 79.83 $\pm$ 0.81 & 88.05 $\pm$ 0.24 & 85.99 $\pm$ 0.34
& \textbf{38.30 $\pm$ 0.89} & 67.05 $\pm$ 0.63 & 59.86 $\pm$ 0.50
& 77.99 & 86.93 & 84.57 \\
\midrule
FedOT(All Global)
& 81.33 $\pm$ 1.19 & 87.30 $\pm$ 0.58 & 85.80 $\pm$ 0.73
& 32.70 $\pm$ 1.02 & 59.20 $\pm$ 3.23 & 52.57 $\pm$ 3.17
& 76.59 & 84.09 & 82.22 \\
FedOT(All Local)
& -- & 88.98 $\pm$ 0.82 & --
& -- & 64.26 $\pm$ 0.66 & --
& -- & 84.54 & -- \\
\midrule
\textsc{FedOT} (Ours)
& \textbf{82.81 $\pm$ 0.47} & \textbf{89.03 $\pm$ 0.21} & \textbf{87.47 $\pm$ 0.26}
& 29.58 $\pm$ 4.52 & 65.43 $\pm$ 3.03 & 56.47 $\pm$ 3.40
& 76.04 & 86.21 & 83.65 \\
\textsc{FedOT}(+B) (Ours)
& \textbf{82.81 $\pm$ 0.47} & \textbf{89.03 $\pm$ 0.21} & \textbf{87.47 $\pm$ 0.26}
& 37.91 $\pm$ 0.90 & \textbf{71.68 $\pm$ 0.61} & \textbf{63.23 $\pm$ 0.44}
& \textbf{78.67} & \textbf{88.58} & \textbf{86.10} \\
\bottomrule\end{tabular}
\end{tabular}
}
\end{table*}

\paragraph{Implications.}
This theorem provides a critical insight: the upper bound on the gradient mismatch between clients is directly proportional to the condition numbers of their respective local transformations. Crucially, if we restrict the local transformations \(w_{\mathrm{l}}^{(i)}\) to be \emph{orthogonal} (which inherently means \(\kappa=1\)), the gradient mismatch among clients is tightly bounded by the constant \(4\tau\). Therefore, orthogonality is not merely beneficial; it is \emph{optimal} for minimizing the upper bound of gradient conflict within this linear transformation framework. This mitigation is essential for facilitating stable global parameter aggregation (enhancing generalization) even under highly heterogeneous (non-IID) data distributions, where gradient directions might otherwise diverge significantly.

\subsection{Benefits of Orthogonal Transformations}
A matrix \(Q\) is orthogonal if \(Q^{\top}Q=I\). All eigenvalues of $Q$ have a magnitude of 1, resulting in the ideal condition number \(\kappa(Q)=1\). Substituting \(w_{\mathrm{l}}^{(i)} = Q\) into \cref{thm:GradDiffBound} yields the minimal mismatch bound \(4\tau\).

\paragraph{Geometric Advantages and Feature Preservation.}
Beyond minimizing gradient conflict, orthogonal transformations offer significant geometric advantages. They are inherently invertible, ensuring that distinct features in the original FM space remain distinct after transformation, thereby preventing information collapse. Furthermore, an orthogonal matrix possesses \(\tfrac{d(d-1)}{2}\) degrees of freedom (DOF), which is approximately half that of a general linear transformation~(\(d^2\)). This constrained capacity is instrumental in preserving the pre-trained manifold structure of the FM, which in turn prevents feature distortion and overfitting to local data.

\subsection{Block-Diagonal Orthogonal Transformations}
\label{sec:block_orthogonal}

We can further regulate the DOF while maintaining  orthogonality via a \emph{block-diagonal} structural constraint.
We partition the dimension \(d\) into \(r\) blocks, with each block \(Q_k\) being independently orthogonal:
$B = \mathrm{diag}\bigl(Q_1,\,\dots,\,Q_r\bigr)$.
Since each block \(Q_k\) is orthogonal, the overall condition number remains \(\kappa(B)=1\).
This structure preserves the lowest bound established in \cref{thm:GradDiffBound} while reducing the total DOF to \(\frac{d\bigl({d}/{r}-1\bigr)}{2}\). This offers a flexible mechanism to trade off between local adaptability and the preservation of global semantic structure.

\section{Experiments}

\label{sec:experiments}
We conduct extensive experiments to evaluate \textsc{FedOT} across two distinct FL scenarios: Cross-Domain FL (stress-testing performance under high heterogeneity) and Cross-Device FL (evaluating scalability in large-scale, heterogeneous environments).

\subsection{Experimental Setup}
\paragraph{Datasets.}
We utilize five diverse datasets characterized by distinct domain shifts: FEMNIST~\cite{RN679} (handwritten digits with writer-specific styles), PACS~\cite{RN774} (variations in image styles), Office-Home~\cite{RN776} (office and home environments), VLCS~\cite{fang2013unbiased} (object recognition across different contexts), and TerraIncognita~\cite{RN777} (wildlife images from disparate locations).

\paragraph{Baselines.}
We compare our proposed methods—\textbf{\textsc{FedOT}} (using a full orthogonal transformation) and its block-diagonal variant \textbf{\textsc{FedOT}(+B)} (detailed in \cref{sec:block_orthogonal})—against a comprehensive suite of baselines. These baselines are grouped as follows:
(1) \textbf{Foundational Benchmark:} CLIP’s zero-shot (ZS) performance.
(2) \textbf{PEFT-FL Methods:} FedCLIP~\cite{RN587} (adapter-based), PromptFL~\cite{guo2023promptfl} (based on CoOp~\cite{zhou2022learning}), CoCoOp~\cite{zhou2022conditional}, and VPT~\cite{jia2022visual}.
(3) \textbf{Adapted PFL Methods:} FedAKT(C) and FedGH(C), which are variants of FedAKT~\cite{liu2025adapter} and FedGH~\cite{yi2023fedgh} modified to utilize the CLIP backbone for an equitable comparison.
(4) \textbf{Architectural Ablations:} Key comparisons to validate our core design choices, including \textbf{FedLT} (using an unconstrained general linear transformation), \textbf{FedAdapter} (using a nonlinear MLP adapter), \textbf{FedOT(All Global)} (all parameters are global), and \textbf{FedOT(All Local)} (all parameters are local). \textbf{FedOT(All Local)} represents a purely local training scenario where no parameters are shared or aggregated. Consequently, a generalized model for unseen domains cannot be formed, rendering the Generalization (G) and Comprehensive (C) metrics non-computable.

\paragraph{Metrics and Implementation.}
We evaluate performance using generalization (G), personalization (P), and comprehensive (C) accuracy, employing a standard leave-one-out cross-domain evaluation protocol~\cite{RN587}.
In this protocol, each domain $i$ is iteratively held out, and we construct an $N \times N$ accuracy matrix $A = [a_{i,j}]$, where $a_{i,j}$ represents a test accuracy of the domain $j$ from the scenario where domain $i$ was excluded.
Specifically, the \textbf{diagonal entry $a_{i,i}$} represents generalization: it is the test accuracy of the aggregated global model when trained on all domains \textit{except} $i$ (i.e., $D_{all} \setminus D_i$), evaluated on the test set of the held-out domain $i$.
The \textbf{off-diagonal entries $a_{i,j}$} (where $j \neq i$) represent personalization: $a_{i,j}$ is the test accuracy of the personalized model of domain $j$ when trained all domains \textit{except} $i$ (i.e., $D_{all} \setminus D_i$), evaluated on the test set of the domain $j$.
From this complete $N \times N$ matrix, we define our metrics:
(1) \textbf{Generalization (G)} quantifies the aggregated performance on unseen held-out domains, defined as the average of the diagonal entries:
$G = \frac{1}{N} \sum_{i=1}^N a_{i,i}$.
(2) \textbf{Personalization (P)} reflects the performance on seen source domains, defined as the average of the off-diagonal entries:
$P = \frac{1}{N(N-1)} \sum_{i=1}^N \sum_{j \neq i} a_{i,j}$.
(3) \textbf{Comprehensive (C)} provides an aggregate measure by averaging all entries:
$C = \frac{1}{N^2} \sum_{i=1}^N \sum_{j=1}^N a_{i,j}$.
We utilize the CLIP ViT-B/32 architecture~\cite{RN568}.
To ensure statistical robustness and reliability, we report the mean and standard deviation across three independent random seeds (50, 77, 98). Detailed implementation settings are provided in Appendix D.
\subsection{Cross-Domain Results and Analysis}

\paragraph{Effectiveness of FedOT.}
As presented in \cref{tab:result}, \textsc{FedOT} and \textsc{FedOT}(+B) consistently outperform nearly all baselines across the evaluation metrics, achieving the highest average comprehensive accuracy (86.10\%). The low standard deviations across multiple seeds confirm the stability and statistical significance of these improvements.
Unlike methods relying solely on global parameters (e.g., FedOT(All Global), FedCLIP), our approach strategically incorporates local parameters, leading to significantly enhanced personalization (Average P=88.58\%) while maintaining competitive generalization (Average G=78.53\%). While FedOT(All Global) achieves the highest generalization on PACS (94.55\%), \textsc{FedOT} delivers superior personalization (97.34\% vs 96.85\%) and comprehensive accuracy (96.64\% vs 96.28\%).
Conversely, the purely local FedOT(All Local) baseline yields an average personalization of 84.54\%. This is substantially outperformed by \textsc{FedOT}(+B) (88.58\%), validating that our federated approach of aggregating global parameters not only enables generalization but also enhances personalization compared to isolated local training.
Notably, methods like FedGH(C) and FedAKT(C) exhibit poor robustness when adapted to the CLIP backbone. This highlights a critical insight: traditional PFL methods designed for lightweight models may not transfer effectively to the FM-FL setting, potentially distorting the FM's inherently well-generalized representations. For comprehensive results, see Appendix F.

\paragraph{Empirical Validation of the Orthogonality Constraint.}
To empirically validate the critical role of the orthogonality constraint ($\kappa=1$), as predicted by \cref{thm:GradDiffBound}, we conduct a detailed analysis focusing on two key aspects: (1) the direct measurement of gradient conflict during training, and (2) the impact of the final condition number ($\kappa$) on generalization accuracy. We compare three distinct architectural choices for the local adaptation module: (1) \textsc{FedOT} (Orthogonal Linear, $\kappa=1$), (2) FedLT (General Linear, $\kappa \ge 1$), and (3) FedAdapter (Nonlinear MLP).

\begin{table}[t]
\caption{\textbf{Analysis of Gradient Conflict during Training.} We report the mean pseudo-gradient cosine similarity ($\mu \pm \sigma$) across all communication rounds (3 seeds, N=12 total runs per dataset). Higher values indicate lower gradient conflict and better alignment. \textsc{FedOT} consistently maintains the highest similarity, validating its superior stability.}
\label{tab:grad_conflict_analysis}
\centering
\resizebox{\columnwidth}{!}{
\begin{tabular}{lccc}
\toprule
\multirow{2}{*}{Dataset} & \textsc{FedOT} & FedLT & FedAdpater\\
 & (Orthogonal, $\kappa=1$) & (General Linear, $\kappa \ge 1$) & (Nonlinear) \\
\midrule\midrule
FEMNIST & \textbf{0.401 $\pm$ 0.013} & 0.309 $\pm$ 0.016 & 0.258 $\pm$ 0.021 \\
VLCS & \textbf{0.002 $\pm$ 0.023} & -0.084 $\pm$ 0.019 & -0.307 $\pm$ 0.048 \\
TerraIncognita & \textbf{0.029 $\pm$ 0.011} & -0.045 $\pm$ 0.011 & -0.235 $\pm$ 0.012 \\
PACS & \textbf{0.036 $\pm$ 0.018} & 0.002 $\pm$ 0.019 & -0.117 $\pm$ 0.052 \\
OfficeHome & \textbf{0.052 $\pm$ 0.011} & 0.000 $\pm$ 0.011 & -0.150 $\pm$ 0.020 \\
\bottomrule
\end{tabular}%
}
\end{table}

\begin{table}[t!]
\caption{\textbf{Impact of Condition Number ($\kappa$) on Accuracy.} 
Comparison of average generalization accuracy (\%) across 3 seeds and the final average condition number $\kappa$. 
Removing the constraint (FedLT) increases $\kappa$ and significantly degrades accuracy, 
especially on highly heterogeneous datasets like FEMNIST.}
\label{tab:condition_accuracy}
\centering
\resizebox{0.9\columnwidth}{!}{
\begin{tabular}{lcccc}
\toprule
\multirow{2}{*}{Dataset} & \multicolumn{2}{c}{\textsc{FedOT} ($\kappa=1$)} & \multicolumn{2}{c}{FedLT ($\kappa \ge 1$)} \\
\cmidrule(lr){2-3}\cmidrule(lr){4-5}
& Avg. Acc. (\%) & $\kappa$ & Avg. Acc. (\%) & Avg $\kappa$ \\
\midrule
\midrule
FEMNIST & \textbf{94.93} & 1.00 & 88.74 & \textbf{32.26} \\
PACS & \textbf{94.53} & 1.00 & 94.32 & 1.24 \\
OfficeHome & \textbf{80.59} & 1.00 & 80.57 & 1.33 \\
VLCS & \textbf{82.81} & 1.00 & 82.21 & 1.24 \\
TerraIncognita & 29.58 & 1.00 & \textbf{29.87} & 1.36 \\
\bottomrule
\end{tabular}
}
\vspace{-0.5em}
\end{table}

First, we analyze gradient conflict directly using the mean pseudo-gradient cosine similarity, summarized in \cref{tab:grad_conflict_analysis}. Supporting our theoretical analysis, \textsc{FedOT} ($\kappa=1$) achieves the highest and most stable average similarity. It reliably holds a near-zero or positive mean value even on highly heterogeneous datasets (e.g., $0.052$ on OfficeHome and $0.036$ on PACS).
By contrast, this alignment breaks down when simply removing the orthogonality constraint (FedLT). Its similarity frequently drops to negative values (e.g., $-0.084$ on VLCS) or hovers near zero ($0.002$ on PACS). The nonlinear FedAdapter baseline performs worst, showing severe and persistent gradient conflict (e.g., $-0.307$ on VLCS and $-0.150$ on OfficeHome).

Second, we demonstrate that this increased gradient conflict directly impacts the final model performance. \cref{tab:condition_accuracy} correlates the average condition number ($\kappa$) at the final training round with the generalization accuracy.
\textsc{FedOT} strictly preserves the condition number ($\kappa = 1.0$) and achieves superior generalization accuracy in most cases.
By contrast, removing the constraint allows the condition number in FedLT to increase substantially, particularly under high heterogeneity (e.g., reaching $32.26$ on FEMNIST). This leads to a marked degradation in generalization performance (dropping from 94.93\% to 88.74\%).
This provides strong empirical confirmation of our theoretical claim (\cref{thm:GradDiffBound}): an unconstrained condition number ($\kappa > 1$) loosens the gradient bound, amplifies gradient discrepancies during training, and ultimately undermines the model's generalization capability.

\begin{figure}[t]
    \centering
    \includegraphics[width=\columnwidth]{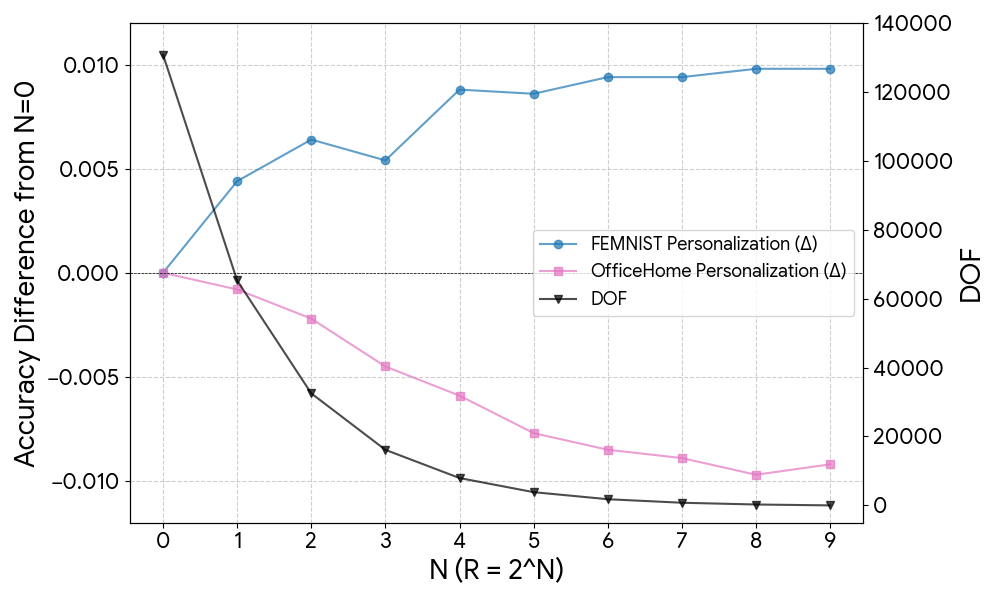}
    \caption{\textbf{Comparison of DOF and personalization accuracy on FEMNIST and OfficeHome.} We vary the number of blocks $R=2^N$. The left axis depicts the change in personalization accuracy relative to $N=0$ ($R=1$). The optimal DOF balances adaptability and structural preservation, varying with dataset complexity.}
    \label{fig:blockwise-personalization-dof}
\end{figure}

\paragraph{DOF and Personalization via FedOT(+B).}
\cref{fig:blockwise-personalization-dof} illustrates the relationship between personalization performance and degrees of freedom (DOF) in \textsc{FedOT}(+B). Simpler tasks (e.g., FEMNIST) benefit from lower DOF, as this constrained capacity helps preserve the pretrained manifold structure and prevents overfitting. Conversely, more complex tasks (e.g., OfficeHome) achieve optimal performance with higher DOF, allowing the model to capture richer, domain-specific patterns. This underscores the necessity of tuning the DOF according to the intrinsic complexity of the dataset.

\subsection{Ablation Study}
\begin{figure}[t]
\centering
\includegraphics[width=\columnwidth]{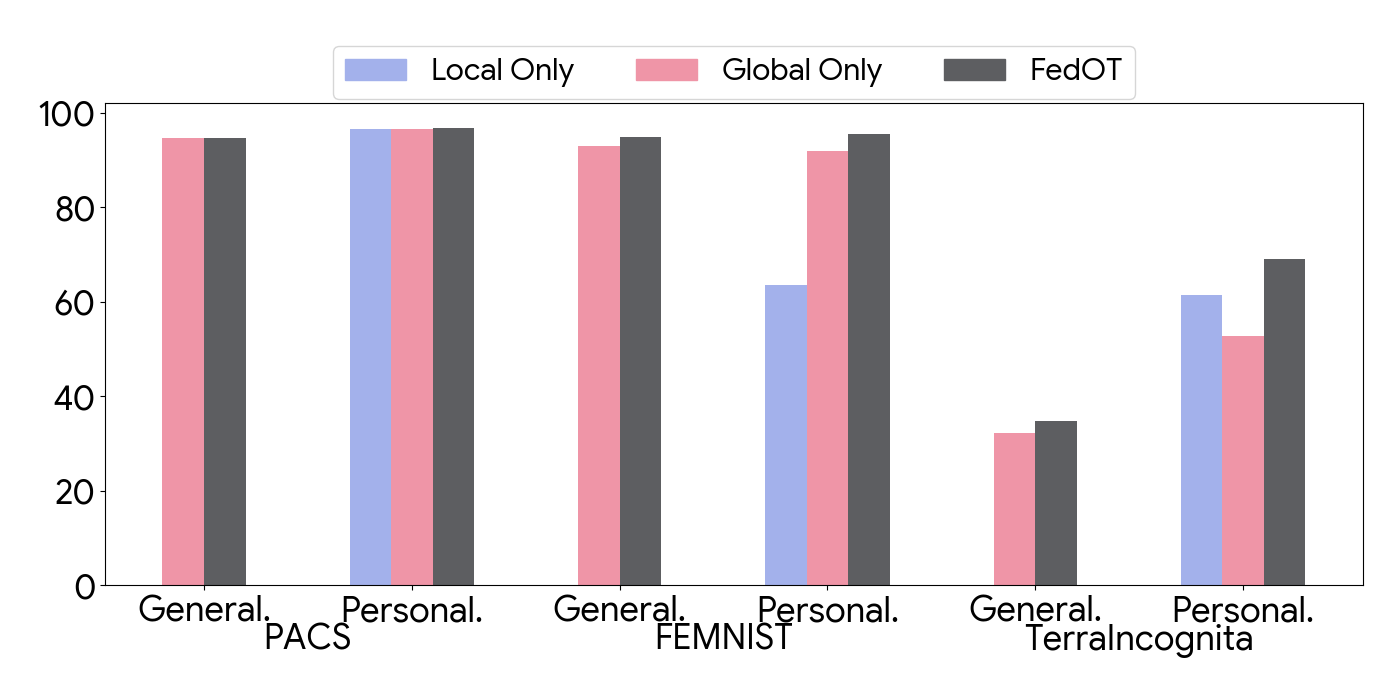}
\caption{\textbf{Comparison of generalization and personalization performance for local-only, global-only, and \textsc{FedOT}.} \textsc{FedOT} consistently outperforms both ablated methods, highlighting the synergistic effect of global and local components.}
\label{fig:global_local}
\end{figure}

\paragraph{Synergy of Global and Local Parameters.}
We compare \textsc{FedOT} with two ablated configurations: one excluding global parameters (local-only) and one excluding local parameters (global-only).
As illustrated in \cref{fig:global_local}, \textsc{FedOT} significantly enhances personalization compared to the local-only approach by leveraging the shared global classifier, which facilitates knowledge transfer across clients. Also, it surpasses the global-only approach in both generalization and personalization by incorporating the client-specific orthogonal transformations, adapting the model to local data distributions.
These results underscore the effectiveness and necessity of jointly integrating both local and global parameters for achieving an optimal balance.

\paragraph{Applicability to Vision-Only Foundation Models.}
To investigate the applicability of our framework beyond VLMs (e.g., CLIP), we conduct experiments using a randomly initialized classifier (RandInit). This simulates a scenario involving a vision-only FM where a text encoder is unavailable for initialization.
As shown in \cref{tab:random_init}, our framework demonstrates highly competitive performance even without VLM pre-alignment.
Notably, on FEMNIST and Office-Home, RandInit actually outperforms initialization using the Text Encoder (TextEnc) (e.g., 96.51\% vs 95.84\% comprehensive accuracy on FEMNIST). This confirms that \textsc{FedOT}'s effectiveness stems primarily from its optimization methodology—orthogonal adaptation and robust aggregation—rather than solely relying on CLIP's inherent multimodal alignment.

\begin{table}[t]
\caption{\textbf{Ablation study on classifier initialization.} Comparison between initialization using CLIP's Text Encoder (TextEnc) vs. Random Initialization (RandInit). RandInit demonstrates competitive performance, confirming \textsc{FedOT}'s applicability to vision-only FMs.}
\label{tab:random_init}
\centering
\resizebox{\columnwidth}{!}{
\begin{tabular}{lcccccc}
\toprule
\multirow{2}{*}{Dataset} & \multicolumn{3}{c}{TextEnc (VLM)} & \multicolumn{3}{c}{RandInit (Vision-only)} \\
\cmidrule(lr){2-4}\cmidrule(lr){5-7}
& G(\%) & P(\%) & C(\%) & G(\%) & P(\%) & C(\%) \\
\midrule
\midrule
FEMNIST & 94.89 & 96.15 & 95.84 & \textbf{96.15} & \textbf{96.63} & \textbf{96.51} \\
PACS & \textbf{94.68} & \textbf{97.44} & \textbf{96.75} & 93.69 & 97.01 & 96.12 \\
Office-Home & 81.30 & 82.92 & 82.51 & \textbf{82.67} & \textbf{83.33} & \textbf{83.17} \\
VLCS & \textbf{74.31} & \textbf{78.07} & \textbf{77.13} & 71.78 & 75.05 & 74.23 \\
TerraIncognita & \textbf{34.80} & \textbf{51.44} & \textbf{47.28} & 34.34 & 49.33 & 45.58 \\
\bottomrule
\end{tabular}
}
\end{table}

\begin{figure}[t]
    \centering
    \includegraphics[width=\columnwidth]{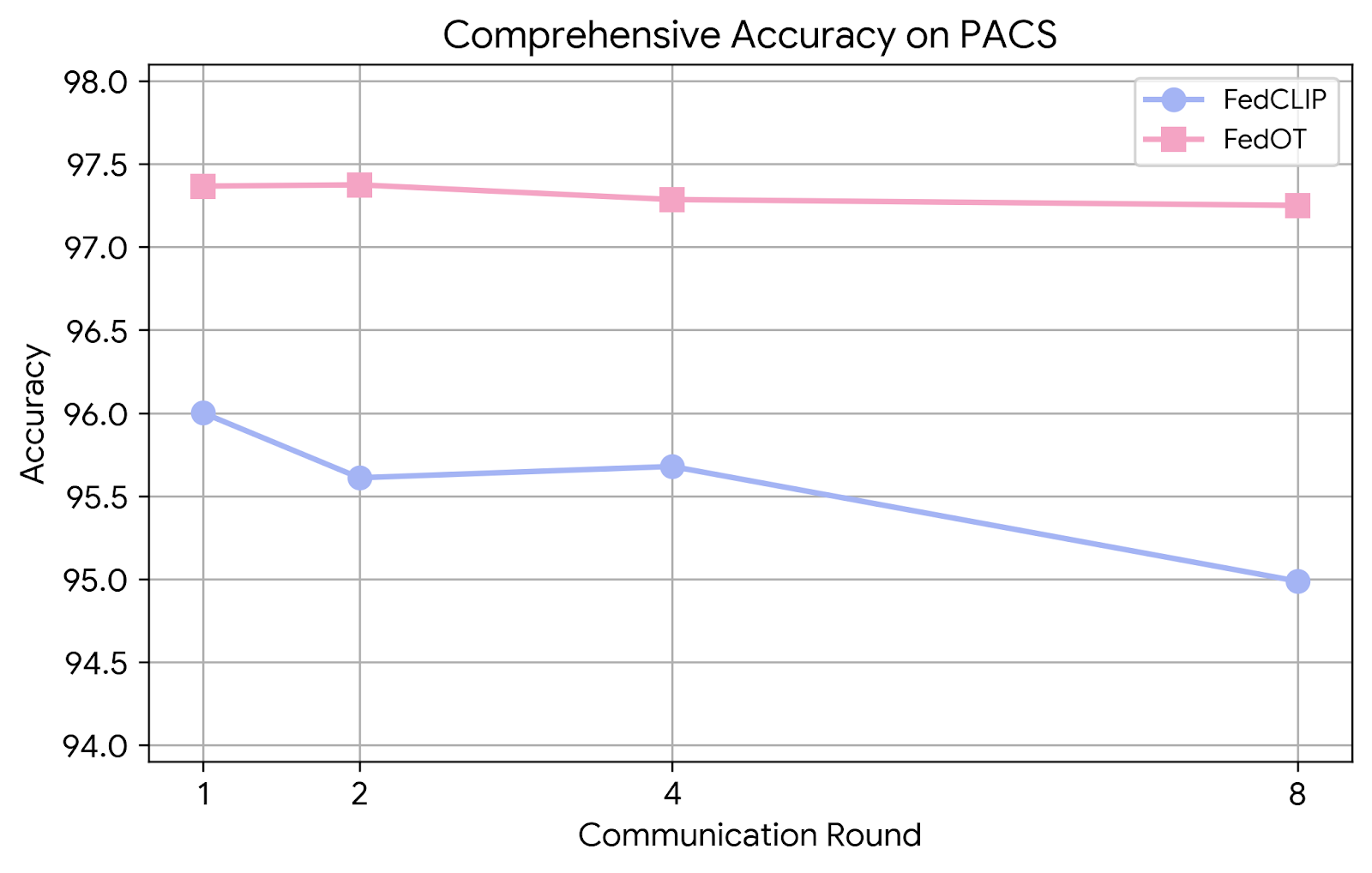}
    \caption{Comparison of comprehensive accuracy on PACS across different communication rounds (varying local epochs E) using \textsc{FedOT} and FedCLIP. \textsc{FedOT} maintains stability and performance, whereas FedCLIP tends to degrade with more communication rounds, highlighting \textsc{FedOT}'s robustness.}
    \label{fig:comm_pacs}
\end{figure}

\paragraph{Robustness to Communication Frequency.}
We assess the impact of varying the local update epoch $E$ (1, 2, 4, 8), effectively reducing the communication frequency. We compare \textsc{FedOT} with FedCLIP on the PACS dataset.
As illustrated in \cref{fig:comm_pacs}, \textsc{FedOT} consistently outperforms FedCLIP and maintains stable performance as the number of communication rounds increases, irrespective of the local epoch setting. This demonstrates the robustness of \textsc{FedOT} to variations in local optimization schedules, a critical feature for practical FL deployments.

\subsection{Cross-Device Scalability Analysis}
\label{sec:scalability}

To assess the applicability of \textsc{FedOT} in a realistic, large-scale cross-device FL environment, we  extended our evaluation using the FEMNIST dataset, which features natural heterogeneity based on user identity. We systematically varied the number of participating training clients (ranging from 1 up to 75) and evaluated the generalization performance on a fixed set of 5 unseen test clients. We compare \textsc{FedOT} against the strongest baseline in this setting, PromptFL.

\begin{figure}[t]
\centering
\includegraphics[width=\columnwidth]{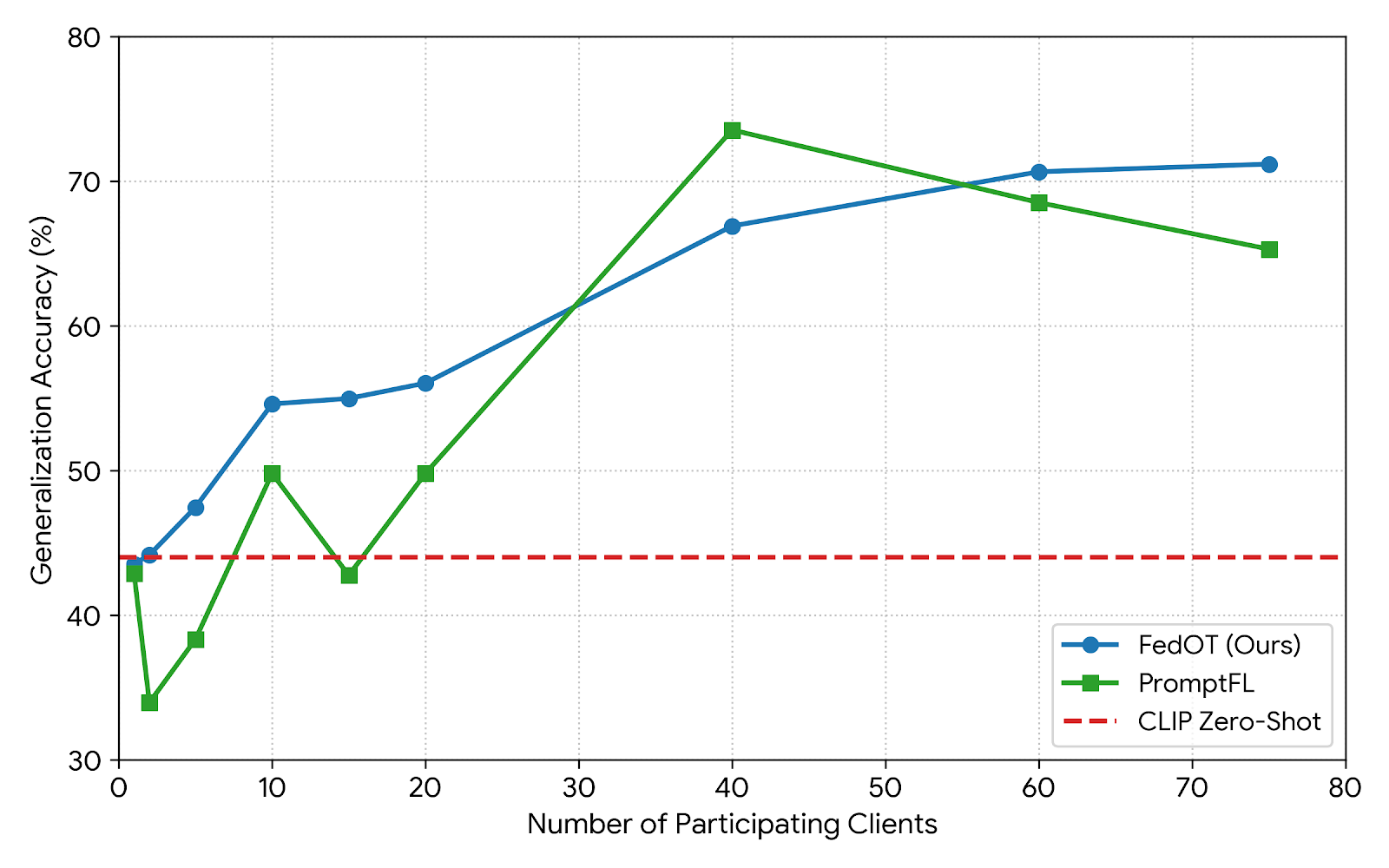}
\caption{\textbf{Scalability analysis on FEMNIST.} Average generalization performance on 5 unseen test clients as the number of participating clients increases from 1 to 75. \textsc{FedOT} consistently improves over the CLIP Zero-Shot baseline and generally outperforms PromptFL, demonstrating robust scalability and stability.}
\label{fig:abl_multi_client}
\end{figure}

As depicted in \cref{fig:abl_multi_client}, the generalization performance of \textsc{FedOT} consistently and monotonically improves as the number of participants increases. With 75 participants, \textsc{FedOT} achieves 71.18\% accuracy, a substantial improvement over the CLIP Zero-Shot (44.00\%).
Furthermore, \textsc{FedOT} generally outperforms PromptFL and demonstrates superior stability. While PromptFL exhibits significant volatility (\eg, dropping sharply from 73.54\% at 40 clients to 65.30\% at 75 clients), \textsc{FedOT} maintains a steady upward trajectory.
These findings confirm that \textsc{FedOT} is highly robust and well-suited for large-scale cross-device FL deployments involving heterogeneous data.

\subsection{Efficiency Analysis}
\label{sec:efficiency}
We analyze the computational overhead and communication cost of \textsc{FedOT}, which are critical factors in practical FL deployments.

\paragraph{Computational Overhead.}
\textsc{FedOT} is computationally efficient because it treats the FM strictly as a black box; gradients are \textit{not} propagated back into the encoder during training. This design is inherently more efficient than methods such as PromptFL, CoCoOp, or VPT, which necessitate full backpropagation through the encoder, even if they possess fewer trainable parameters.
As analyzed in Tab. 7
(Appendix), \textsc{FedOT}'s per-epoch training time is  competitive. Crucially, its computation cost remains fixed regardless of the underlying backbone complexity, ensuring excellent scalability as FMs continue to increase in size.

\paragraph{Communication Cost.}
\textsc{FedOT} is also communication-efficient. Only the parameters of the global classifier ($K \times d$) are exchanged between the clients and the server. The local orthogonal transformation parameters ($d \times d$) remain strictly on-device.
The total number of communicated parameters is remarkably small (\eg, $\sim$5K parameters for FEMNIST), which is orders of magnitude smaller than traditional FL methods that transmit full models or large gradient vectors. 

\section{Limitations and Future Works}
\label{sec:limit_future}
Although \textsc{FedOT} treats the FM encoder as a black box, the utilization of a shared global classifier introduces some white-box elements (the classifier weights themselves), which may be restrictive in scenarios where only inference API queries are permitted (e.g., fully black-box FL).\\
Future research may explore the theoretical integration of nonlinear mappings—instead of assumption of linear transformations—to potentially build a new insight. While this work focused on vision tasks, the modality-agnostic nature of the \textsc{FedOT} framework suggests promising extensions to Large Language Models (LLMs) by applying orthogonal transformations to token embeddings, a direction supported by related findings in the OFT literature~\cite{RN683}.

\section{Conclusion}
We introduced \textsc{FedOT}, a federated learning framework specifically designed to address the practical constraints of utilizing black-box foundation models while ensuring dual privacy. By integrating a globally shared classifier with local orthogonal transformations applied externally to the FM embeddings, \textsc{FedOT} effectively balances between generalization and personalization. Our theoretical analysis proves that the orthogonality constraint minimizes the upper bound of gradient conflicts across clients. Extensive empirical validation across diverse benchmarks demonstrates consistent performance improvements and robust scalability. \textsc{FedOT} provides a practical, efficient, and secure solution for deploying real-world federated applications that leverage proprietary foundation models across diverse domains.

\section*{Acknowledgment}
This work was supported by the National Research Foundation of Korea (NRF) grant funded by the Korea government (MSIT) (No. RS-2024-00345809, Research on AI Robustness Against Distribution Shift in Real-World Scenarios; and No. RS-2023-00222663, Center for Optimizing Hyperscale AI Models and Platforms).

{
    \small
    \bibliographystyle{ieeenat_fullname}
    \bibliography{main}
}


\end{document}


\maketitle
\thispagestyle{empty}

\appendix

\section{Related Work (Extended)}
\label{appendix:related}
\subsection{Foundation Models}
Foundation models~\cite{Bommasani2021FM}, exemplified by the GPT series~\cite{RN744,RN745,RN746} and VLMs\cite{RN568, jia2021scaling}, are characterized by their substantial scale and extensive pre-training on broad data. CLIP~\cite{RN568} notably showcases the potential of aligning visual and language modalities through contrastive learning. In this study, we concentrate on vision foundation models, including VLMs, and highlight their significant potential for extension to federated learning paradigms.

\subsection{Federated Learning}
FL facilitates collaborative model training across decentralized data sources~\cite{mcmahan2017communication}. Heterogeneous FL specifically addresses challenges arising from non-IID data distributions, which complicate model aggregation and convergence~\cite{RN649}. Personalized FL (PFL) aims to mitigate these challenges by training personalized local models alongside a generalized global model, adapting to local heterogeneity~\cite{pillutla2022federated, RN723}. The integration of large-scale pre-trained VLMs within the FL framework represents an active and promising area of research~\cite{RN587,guo2023promptfl}.

\section{FedOT Algorithm Details}

\subsection{Pseudo-code}
The detailed algorithm of \textsc{FedOT} is presented in the form of a complete pseudo-code in \cref{tab:alg1}.

\begin{algorithm}[H]
\caption{\textbf{\textsc{FedOT}}: Orthogonal Local Updates via Cayley Transform}
\label{tab:alg1}
\begin{algorithmic}[1]
\Statex \textbf{Global initialisation}
\State Initialise server parameter $w_{\mathrm{g},0}$
\ForAll{client $i\in\{1,\dots,N\}$ \textbf{in parallel}}
    \State $X^{(i)}_0 \gets I$
    \State $P^{(i)}_0 \gets \tfrac12\!\bigl(X^{(i)}_0 - X^{(i)\top}_0\bigr)$
    \State $w^{(i)}_{\mathrm{l},0} \gets (I + P^{(i)}_0)(I - P^{(i)}_0)^{-1}$
\EndFor
\State Server broadcasts $w_{\mathrm{g},0}$ to all clients
\vspace{0.4em}

\For{round $t = 0$ \textbf{to} $T-1$}
    \ForAll{client $i$ \textbf{in parallel}}
        \State $(w^{(i)}_{\mathrm{g},t+1},\,X^{(i)}_{t+1})
             \gets \Call{LocalUpdate}{i,\,w_{\mathrm{g},t},\,X^{(i)}_t}$
        \State Client sends $w^{(i)}_{\mathrm{g},t+1}$ to server
    \EndFor
    \Statex \textbf{Server aggregation}
    \State $w_{\mathrm{g},t+1} \gets \frac1N\sum_{i=1}^N w^{(i)}_{\mathrm{g},t+1}$
    \State Server broadcasts $w_{\mathrm{g},t+1}$ to all clients
\EndFor
\vspace{0.4em}

\Function{LocalUpdate}{$i,\,w_{\mathrm{g}},\,X^{(i)}$}
    \State $P^{(i)} \gets \tfrac12\!\bigl(X^{(i)} - X^{(i)\top}\bigr)$
    \State $w^{(i)}_{\mathrm{l}} \gets (I + P^{(i)})(I - P^{(i)})^{-1}$
    \For{epoch $e = 1$ \textbf{to} $E$}
        \ForAll{mini-batch $B \subset D^{(i)}$}
            \State $w^{(i)}_{\mathrm{g}}\! \gets w^{(i)}_{\mathrm{g}}
                   - \eta\,\nabla_{w^{(i)}_{\mathrm{g}}}\ell^{(i)}(w^{(i)}_{\mathrm{g}},\,w^{(i)}_{\mathrm{l}};B)$
            \State $X^{(i)} \gets X^{(i)}
                   - \eta\,\nabla_{X^{(i)}}\ell^{(i)}(w^{(i)}_{\mathrm{g}},\,w^{(i)}_{\mathrm{l}};B)$
            \State $P^{(i)} \gets \tfrac12\!\bigl(X^{(i)} - X^{(i)\top}\bigr)$
            \State $w^{(i)}_{\mathrm{l}} \gets (I + P^{(i)})(I - P^{(i)})^{-1}$ \Comment{Cayley transform}
        \EndFor
    \EndFor
    \State \Return $(w^{(i)}_{\mathrm{g}},\,X^{(i)})$ \Comment{Only $w^{(i)}_{\mathrm{g}}$ sent to server}
\EndFunction
\end{algorithmic}
\end{algorithm}

\subsection{Cayley Transform}
We utilize the Cayley transform~\cite{cayley1846quelques} to enforce strict orthogonality during optimization. The orthogonal matrix $Q$ is derived using $Q = (I + P)(I - P)^{-1}$, where $P$ is a skew-symmetric matrix ($P = -P^\top$). We formulate $P$ as $P = 0.5(X - X^\top)$, where $X$ is an unconstrained transformation matrix. Because $P$ is skew-symmetric, $(I - P)$ is guaranteed to be invertible, ensuring the Cayley transform is well-defined and continuously differentiable. This allows for stable, standard gradient-based optimization directly on the unconstrained matrix $X$.

\section{Theoretical Analysis (Proofs)}
\label{sec:theoretical analysis}

\subsection{Proof of Theorem 1 (Bounding the Gradient Difference)}
\label{sec:bounding_gdiff}

We define the cross-entropy loss for client $i$: 
$\ell^{(i)} = \mathbb{E}_{(x,y)\in D^{(i)}} \bigl[\mathrm{CE}(r_{x}^{(i)}, y)\bigr]$.
The gradient with respect to the global classifier $w_{\mathrm{g}}^{(i)}$ is:
\[
  \nabla_{w_{\mathrm{g}}^{(i)}} \ell^{(i)}
  = \mathbb{E}\Bigl[
      \tau \,\bigl(r_{x}^{(i)} - y\bigr)\,
      \Bigl(\frac{w_{\mathrm{l}}^{(i)}\,\mathcal{I}(x)}
                  {\|w_{\mathrm{l}}^{(i)}\,\mathcal{I}(x)\|}
      \Bigr)^{\!\top}
    \Bigr].
\]

Let $\Delta_i := r_{x}^{(i)} - y$.
Since $r_{x}^{(i)}$ is a probability distribution vector and $y$ is a one-hot vector,
we know $\|\Delta_i\| \le 2$ (derived from the triangle inequality: $\|r_x^{(i)}\| \le 1$ and $\|y\| \le 1$).

For clients $i$ and $j$, we analyze the difference between their gradients using the triangle inequality:
\begin{align}
&\Bigl\|\nabla_{w_{\mathrm{g}}^{(i)}} \ell^{(i)}
       - \nabla_{w_{\mathrm{g}}^{(j)}} \ell^{(j)}\Bigr\|_{2}
\nonumber\\
&\quad\le\;
  \mathbb{E}_{D^{(i)}}
    \Bigl\|\tau\,\Delta_i\,
      \frac{w_{\mathrm{l}}^{(i)}\,\mathcal{I}(x)}
           {\|w_{\mathrm{l}}^{(i)}\,\mathcal{I}(x)\|}
    \Bigr\|
  \;+\;
  \mathbb{E}_{D^{(j)}}
    \Bigl\|\tau\,\Delta_j\,
      \frac{w_{\mathrm{l}}^{(j)}\,\mathcal{I}(x)}
           {\|w_{\mathrm{l}}^{(j)}\,\mathcal{I}(x)\|}
    \Bigr\|
\nonumber\\
&\quad\le\;
  \mathbb{E}_{D^{(i)}}
    \Bigl\|2\tau\,
      \frac{w_{\mathrm{l}}^{(i)}\,\mathcal{I}(x)}
           {\|w_{\mathrm{l}}^{(i)}\,\mathcal{I}(x)\|}
    \Bigr\|
  \;+\;
  \mathbb{E}_{D^{(j)}}
    \Bigl\|2\tau\,
      \frac{w_{\mathrm{l}}^{(j)}\,\mathcal{I}(x)}
           {\|w_{\mathrm{l}}^{(j)}\,\mathcal{I}(x)\|}
    \Bigr\|.
\label{eq:grad_bound_step}
\end{align}

By definition, the norm of the transformed normalized vector is bounded by the condition number $\kappa$ of the transformation matrix:
\[
  \frac{\|w_{\mathrm{l}}^{(i)}\,\mathcal{I}(x)\|}
       {\|w_{\mathrm{l}}^{(i)}\,\mathcal{I}(x)\|}
  \le \kappa\bigl(w_{\mathrm{l}}^{(i)}\bigr),
\]
which leads to the final bound:
\[
  \Bigl\|\nabla_{w_{\mathrm{g}}^{(i)}} \ell^{(i)}
        - \nabla_{w_{\mathrm{g}}^{(j)}} \ell^{(j)}\Bigr\|
  \;\le\; 2\tau\,\Bigl[\kappa\bigl(w_{\mathrm{l}}^{(i)}\bigr)
                       + \kappa\bigl(w_{\mathrm{l}}^{(j)}\bigr)\Bigr].
\]

\subsection{Condition Number of an Orthogonal Transformation}
An orthogonal transformation \( Q \) satisfies \( Q^\top Q = I \). Consider an eigenvector $v$ such that \( Qv = \lambda v \). Taking the inner product:
$\langle Qv, Qv \rangle = |\lambda|^2 \langle v, v \rangle$.
Alternatively, using the property of orthogonality: $\langle Qv, Qv \rangle = \langle v, Q^\top Q v \rangle = \langle v, I v \rangle = \langle v, v \rangle$.
Equating both expressions yields \( |\lambda|^2 = 1 \). Since the singular values of $Q$ are the square roots of the eigenvalues of $Q^\top Q = I$, all singular values are exactly 1. Therefore, the condition number, defined as the ratio of the largest to the smallest singular value, is fixed at \( \kappa(Q) = 1/1 = 1 \).

\section{Experimental Setup Details} \label{appendix:setup}

\subsection{Datasets}
We utilize five distinct datasets for our evaluation. FEMNIST~\cite{RN679} utilizes digit samples from four specific users (IDs: 0, 25, 26, 33) to simulate cross-domain heterogeneity based on writing style. PACS~\cite{RN774} comprises four domains characterized by stylistic differences (Photo, Art, Cartoon, Sketch). VLCS~\cite{fang2013unbiased} consists of four sub-datasets representing different contexts (VOC2007, LabelMe, Caltech101, SUN09). Office-Home~\cite{RN776} includes four domains with varying visual appearances (Art, Clipart, Product, Real-world). TerraIncognita~\cite{RN777} utilizes images captured at four distinct geographical locations (L38, L43, L46, L100). For all datasets, we employ a standardized 60/20/20 split for training, validation, and testing sets.

\subsection{Implementation Details}
We utilize the CLIP pre-trained model with the ViT-B/32 architecture~\cite{RN568}. The local update epoch is set to $E=1$. The total number of communication rounds is $T=200$ for PACS, FEMNIST, and VLCS, and $T=50$ for Office-Home and TerraIncognita, reflecting differences in dataset complexity and convergence speed.
We use the SGD optimizer with a weight decay of $5e-4$. The learning rates are tuned for optimal performance: $1 \times 10^{-3}$ for FEMNIST, and $5 \times 10^{-5}$ for the other datasets. The optimal block size for \textsc{FedOT(+B)} for each dataset are as follows: 256 for FEMNIST, 1 for PACS, OfficeHome, and VLCS, and 4 for TerraIncognita. 

\subsection{Computational Resources and Efficiency Analysis}
Experiments were conducted on high-performance computing clusters equipped with NVIDIA A6000 (48GB) and RTX 3090 (24GB) GPUs, utilizing the PyTorch 2.1.0 framework~\cite{RN145}. We provide a detailed analysis of runtime and computational overhead across representative FL baselines on the FEMNIST dataset in \cref{tab:computation_analysis}.

\clearpage

\onecolumn

\section{Additional Experimental Results}

\subsection{Detailed Cross-Device Scalability Results}
\label{appendix:scalability_details}
For the scalability experiments detailed in Sec. 5.4, we utilized the FEMNIST dataset. To ensure stable training conditions while preserving the inherent heterogeneity (writer-specific data partitions), we pre-filtered the dataset to retain only users possessing a sufficient number of data samples (at least 100 images) and randomly sampled participants from this refined pool. This minimum-samples-per-client criterion was employed solely to isolate the effect of the number of participants on performance and does not alter the underlying feature heterogeneity. We designated a fixed set of 5 unseen test clients (User IDs: 1894, 224, 3321, 745, 509) and systematically varied the number of participating clients (from 1 to 75) per communication round.

\cref{tab:scalability_fedot} presents the detailed generalization results for \textsc{FedOT}, and \cref{tab:scalability_promptfl} presents the corresponding results for the PromptFL baseline.

\begin{table*}[h]
\centering
\caption{Detailed Scalability Results for \textsc{FedOT} on FEMNIST. Generalization accuracy (\%) on 5 unseen test clients.}
\label{tab:scalability_fedot}
\begin{tabular}{ccccccc}
\toprule
\multirow{2}{*}{No. of Participants} & \multicolumn{5}{c}{Test Clients (User ID)} & \multirow{2}{*}{AVG} \\
\cmidrule(lr){2-6}
& \textbf{1894} & \textbf{224} & \textbf{3321} & \textbf{745} & \textbf{509} & \\
\midrule \midrule
0 (=CLIP ZS) & 33.33\% & 26.32\% & 44.00\% & 68.75\% & 47.62\% & 44.00\% \\
1 & 29.17\% & 42.11\% & 44.00\% & 50.00\% & 52.38\% & 43.53\% \\
2 & 33.33\% & 31.58\% & 52.00\% & 56.25\% & 47.62\% & 44.16\% \\
5 & 33.33\% & 31.58\% & 56.00\% & 68.75\% & 47.62\% & 47.46\% \\
10 & 37.50\% & 31.58\% & 56.00\% & 81.25\% & 66.67\% & 54.60\% \\
15 & 41.67\% & 31.58\% & 60.00\% & 75.00\% & 66.67\% & 54.98\% \\
20 & 41.67\% & 36.84\% & 60.00\% & 75.00\% & 66.67\% & 56.04\% \\
40 & 54.17\% & 57.89\% & 76.00\% & 75.00\% & 71.43\% & 66.90\% \\
60 & 66.67\% & 57.89\% & 76.00\% & 81.25\% & 71.43\% & 70.65\% \\
75 & 70.83\% & 57.89\% & 76.00\% & 75.00\% & 76.19\% & 71.18\% \\
\bottomrule
\end{tabular}
\end{table*}

\begin{table*}[h]
\centering
\caption{Detailed Scalability Results for PromptFL on FEMNIST. Generalization accuracy (\%) on 5 unseen test clients.}
\label{tab:scalability_promptfl}
\begin{tabular}{ccccccc}
\toprule
\multirow{2}{*}{No. of Participants} & \multicolumn{5}{c}{Test Clients (User ID)} & \multirow{2}{*}{AVG} \\
\cmidrule(lr){2-6}
& \textbf{1894} & \textbf{224} & \textbf{3321} & \textbf{745} & \textbf{509} & \\
\midrule \midrule
1 & 33.33\% & 26.32\% & 48.00\% & 68.75\% & 38.10\% & 42.90\% \\
2 & 29.17\% & 21.05\% & 44.00\% & 37.50\% & 38.10\% & 33.96\% \\
5 & 25.00\% & 21.05\% & 48.00\% & 50.00\% & 47.62\% & 38.33\% \\
10 & 37.50\% & 26.32\% & 56.00\% & 62.50\% & 66.67\% & 49.80\% \\
15 & 33.33\% & 31.58\% & 48.00\% & 43.75\% & 57.14\% & 42.76\% \\
20 & 37.50\% & 26.32\% & 56.00\% & 62.50\% & 66.67\% & 49.80\% \\
40 & 58.33\% & 47.37\% & 84.00\% & 87.50\% & 90.48\% & 73.54\% \\
60 & 50.00\% & 47.37\% & 72.00\% & 87.50\% & 85.71\% & 68.52\% \\
75 & 54.17\% & 42.11\% & 68.00\% & 81.25\% & 80.95\% & 65.30\% \\
\bottomrule
\end{tabular}
\end{table*}

\clearpage

\subsection{Computational Analysis}
We conduct a computational analysis on FEMNIST. As depicted on the Table~\ref{tab:computation_analysis}, our FedOT gives relatively comparable runtime to other methods, and also in parameters.
\begin{table*}[h] 
\centering
\caption{Computational Analysis on FEMNIST. $d_{txtp}, d_{imgp}$: prompt dimensions; $d_{txtemb}, d_{imgemb}$: embedding dimensions; $K$: number of classes; $n_{ctx}$: context length. Meta-net size for CoCoOp is calculated as $d_{imgemb}^2/16 + d_{imgemb}/16 + d_{imgemb}/16 \times d_{txtp} + d_{txtp}$.}
\label{tab:computation_analysis}
\begin{tabular}{@{}lllc@{}} 
\toprule
Method & Runtime & Type & Parameters \\
\midrule \midrule
\multirow{2}{*}{PromptFL} & \multirow{2}{*}{22s} & Server & $d_{txtp} \times n_{ctx}$ \\
 & & Client & $d_{txtp} \times n_{ctx}$ \\
\cmidrule(l){1-4} 
\multirow{2}{*}{CoCoOp} & \multirow{2}{*}{2m 9s} & Server & $d_{txtp} \times n_{ctx}$ \\
 & & Client & $d_{txtp} \times n_{ctx} + \text{meta-net}$ \\
\cmidrule(l){1-4}
\multirow{2}{*}{VPT} & \multirow{2}{*}{26s} & Server & $K \times d_{txtemb}$ \\
 & & Client & $K \times d_{txtemb} + d_{imgp} \times n_{ctx}$ \\
\cmidrule(l){1-4}
\multirow{2}{*}{FedAKT(C)} & \multirow{2}{*}{23s} & Server & $2 \times (d_{imgemb}^2 + d_{imgemb})$ \\
 & & Client & $K \times d_{txtemb} + K$ \\
\cmidrule(l){1-4}
\multirow{2}{*}{FedGH(C)} & \multirow{2}{*}{27s} & Server & $K \times d_{txtemb} + K$ \\
 & & Client & $K \times d_{txtemb} + K$ \\
\cmidrule(l){1-4}
\multirow{2}{*}{FedCLIP} & \multirow{2}{*}{19s} & Server & $2 \times (d_{imgemb}^2 + d_{imgemb})$ \\
 & & Client & $2 \times (d_{imgemb}^2 + d_{imgemb})$ \\
\cmidrule(l){1-4}
\multirow{2}{*}{FedOT(All Global)} & \multirow{2}{*}{28s} & Server & $K \times d_{txtemb} + d_{imgemb}^2$ \\
 & & Client & $K \times d_{txtemb} + d_{imgemb}^2$ \\
\cmidrule(l){1-4}
\multirow{2}{*}{\textsc{FedOT} (Ours)} & \multirow{2}{*}{26s} & Server & $K \times d_{txtemb}$ \\
 & & Client & $K \times d_{txtemb} + d_{imgemb}^2$ \\
\cmidrule(l){1-4}
\multirow{2}{*}{\textsc{FedOT}(+B) (Ours)} & \multirow{2}{*}{24s} & Server & $K \times d_{txtemb}$ \\
 & & Client & $K \times d_{txtemb} + d_{imgemb}^2$ \\
\bottomrule
\end{tabular}
\end{table*}

\clearpage
\section{Full Results of Main Experiment} \label{appendix:full_results}
\subsection{Generalization}
\begin{table}[htb!]
\centering
\resizebox{\columnwidth}{!}{%
\begin{tabular}{l|cccc|c}
\toprule
Method & 33 & 0 & 25 & 26 & AVG \\ 
\midrule\midrule
Zero Shot 
& $50.88 \pm 3.42$ & $48.40 \pm 1.47$ & $45.19 \pm 2.50$ & $48.24 \pm 2.73$ & $48.18 \pm 1.31$ \\
FedCLIP
& $68.57 \pm 8.25$ & $70.99 \pm 5.57$ & $72.12 \pm 12.01$ & $70.51 \pm 9.15$ & $70.55 \pm 8.60$ \\
PromptFL ($M=4$)
& $94.28 \pm 1.78$ & $96.63 \pm 0.49$ & $96.80 \pm 1.00$ & $97.44 \pm 0.74$ & $96.29 \pm 0.86$ \\
PromptFL ($M=16$)
& $95.51 \pm 0.41$ & $96.96 \pm 0.28$ & $96.64 \pm 0.84$ & \textbf{97.76 $\pm$ 0.28} & $96.72 \pm 0.30$ \\
CoCoOp (All Global)
& $86.26 \pm 10.82$ & $87.02 \pm 9.47$ & $87.50 \pm 7.51$ & $87.34 \pm 7.73$ & $87.03 \pm 8.86$ \\
CoCoOp (Meta-Net Random Init)
& $21.22 \pm 12.57$ & $20.67 \pm 11.81$ & $23.56 \pm 10.84$ & $24.04 \pm 12.91$ & $22.37 \pm 11.79$ \\
CoCoOp (Server CoOp)
& $15.10 \pm 1.23$ & $17.31 \pm 5.42$ & $17.63 \pm 9.50$ & $9.62 \pm 2.89$ & $14.91 \pm 3.27$ \\
\bottomrule
\end{tabular}%
}
\caption{The generalization accuracy of FEMNIST.}
\label{tab:gen_femnist}
\end{table}

\begin{table}[htb!]
\centering
\resizebox{\columnwidth}{!}{%
\begin{tabular}{l|cccc|c}
\toprule
Method & A & C & P & S & AVG \\ 
\midrule\midrule
Zero Shot 
& $95.35 \pm 0.98$ & $96.94 \pm 0.81$ & $99.60 \pm 0.46$ & $85.56 \pm 0.99$ & $94.36 \pm 0.48$ \\
FedCLIP
& $95.36 \pm 1.36$ & $96.58 \pm 0.57$ & $99.30 \pm 0.46$ & $85.73 \pm 1.71$ & $94.24 \pm 0.71$ \\
PromptFL ($M=4$)
& $95.27 \pm 0.75$ & $97.22 \pm 0.57$ & $99.50 \pm 0.46$ & $84.25 \pm 1.95$ & $94.06 \pm 0.61$ \\
PromptFL ($M=16$)
& $94.05 \pm 1.10$ & $97.15 \pm 0.65$ & $99.40 \pm 0.30$ & $85.94 \pm 1.41$ & $94.14 \pm 0.57$ \\
CoCoOp (All Global)
& $95.27 \pm 2.19$ & $97.08 \pm 1.05$ & $98.90 \pm 0.62$ & $85.69 \pm 0.96$ & $94.24 \pm 1.11$ \\
CoCoOp (Meta-Net Random Init)
& $62.10 \pm 13.94$ & $77.35 \pm 8.04$ & $62.77 \pm 15.22$ & $72.57 \pm 8.93$ & $68.70 \pm 11.29$ \\
CoCoOp (Server CoOp)
& $85.74 \pm 0.61$ & $74.58 \pm 25.97$ & $89.42 \pm 6.81$ & $80.21 \pm 11.88$ & $82.49 \pm 8.16$ \\
\bottomrule
\end{tabular}%
}
\caption{The generalization accuracy of PACS.}
\label{tab:gen_pacs}
\end{table}

\begin{table}[htb!]
\centering
\resizebox{\columnwidth}{!}{%
\begin{tabular}{l|cccc|c}
\toprule
Method & A & C & P & R & AVG \\ 
\midrule\midrule
Zero Shot
& $77.59 \pm 1.33$ & $65.29 \pm 0.89$ & $86.89 \pm 1.61$ & $86.83 \pm 0.74$ & $79.15 \pm 0.40$ \\
FedCLIP
& $77.32 \pm 1.07$ & $65.79 \pm 1.56$ & $85.94 \pm 1.42$ & $86.68 \pm 1.99$ & $78.93 \pm 1.06$ \\
PromptFL ($M=4$)
& $77.39 \pm 2.00$ & $68.12 \pm 1.00$ & $87.86 \pm 0.51$ & $88.33 \pm 0.86$ & $80.42 \pm 0.80$ \\
PromptFL ($M=16$)
& \textbf{79.86 $\pm$ 0.43} & $68.50 \pm 1.60$ & $88.47 \pm 0.46$ & \textbf{89.13 $\pm$ 0.88} & \textbf{81.49 $\pm$ 0.45} \\
CoCoOp (All Global)
& $77.73 \pm 1.89$ & $65.56 \pm 2.83$ & $88.39 \pm 0.85$ & $88.29 \pm 1.20$ & $79.99 \pm 0.90$ \\
CoCoOp (Meta-Net Random Init)
& $60.96 \pm 8.08$ & $50.13 \pm 2.75$ & $67.53 \pm 4.39$ & $69.00 \pm 4.67$ & $61.91 \pm 4.84$ \\
CoCoOp (Server CoOp)
& $72.00 \pm 1.46$ & $73.00 \pm 1.46$ & $73.50 \pm 1.46$ & $72.42 \pm 1.46$ & $72.73 \pm 1.46$ \\
\bottomrule
\end{tabular}%
}
\caption{The generalization accuracy of Office-Home.}
\label{tab:gen_oh}
\end{table}

\begin{table}[htb!]
\centering
\resizebox{\columnwidth}{!}{%
\begin{tabular}{l|cccc|c}
\toprule
Method & C & L & S & V & AVG \\ 
\midrule\midrule
Zero Shot
& \textbf{100.00 $\pm$ 0.00} & \textbf{65.22 $\pm$ 2.72} & $70.96 \pm 0.33$ & $84.79 \pm 1.07$ & $80.59 \pm 0.96$ \\
FedCLIP
& $99.29 \pm 0.71$ & $63.40 \pm 0.54$ & $73.48 \pm 1.40$ & $85.63 \pm 1.71$ & $80.45 \pm 0.86$ \\
PromptFL ($M=4$)
& $99.06 \pm 0.54$ & $62.96 \pm 1.52$ & $78.96 \pm 1.30$ & $86.17 \pm 1.56$ & $81.79 \pm 0.15$ \\
PromptFL ($M=16$)
& $99.18 \pm 0.54$ & $60.20 \pm 1.21$ & $77.75 \pm 2.51$ & $83.90 \pm 2.20$ & $80.26 \pm 0.77$ \\
CoCoOp (All Global)
& $99.88 \pm 0.20$ & $62.65 \pm 4.18$ & $74.34 \pm 4.05$ & $84.05 \pm 4.20$ & $80.23 \pm 1.96$ \\
CoCoOp (Meta-Net Random Init)
& $67.02 \pm 9.08$ & $49.46 \pm 6.01$ & $52.75 \pm 9.02$ & $55.41 \pm 7.63$ & $56.16 \pm 6.06$ \\
CoCoOp (Server CoOp)
& $89.28 \pm 18.57$ & $53.99 \pm 12.15$ & $60.36 \pm 3.58$ & $74.77 \pm 0.17$ & $69.60 \pm 7.67$ \\
\bottomrule
\end{tabular}%
}
\caption{The generalization accuracy of VLCS.}
\label{tab:gen_vlcs}
\end{table}

\begin{table}[htb!]
\centering
\resizebox{\columnwidth}{!}{%
\begin{tabular}{l|cccc|c}
\toprule
Method & L38 & L43 & L46 & L100 & AVG \\ 
\midrule\midrule
Zero Shot
& $13.35 \pm 0.76$ & $29.90 \pm 1.95$ & $19.59 \pm 1.31$ & $13.47 \pm 0.52$ & $19.08 \pm 0.26$ \\
FedCLIP
& $33.28 \pm 4.81$ & $30.98 \pm 3.84$ & $30.67 \pm 1.80$ & \textbf{43.56 $\pm$ 9.00} & $34.62 \pm 1.63$ \\
PromptFL ($M=4$)
& $38.54 \pm 2.98$ & $33.02 \pm 2.20$ & $36.99 \pm 2.26$ & $41.47 \pm 4.42$ & $37.51 \pm 1.92$ \\
PromptFL ($M=16$)
& $37.31 \pm 7.34$ & $34.43 \pm 3.48$ & $35.28 \pm 3.40$ & $36.96 \pm 1.72$ & $35.99 \pm 2.50$ \\
CoCoOp (All Global)
& $38.33 \pm 7.94$ & $29.82 \pm 2.76$ & $38.84 \pm 1.38$ & $37.61 \pm 1.36$ & $36.15 \pm 1.44$ \\
CoCoOp (Meta-Net Random Init)
& $44.65 \pm 2.62$ & $26.09 \pm 5.39$ & $25.52 \pm 4.81$ & $23.18 \pm 2.98$ & $29.86 \pm 1.21$ \\
CoCoOp (Server CoOp)
& $34.15 \pm 18.23$ & $26.34 \pm 5.58$ & $24.16 \pm 5.63$ & $20.03 \pm 3.40$ & $26.17 \pm 5.38$ \\
\bottomrule
\end{tabular}%
}
\caption{The generalization accuracy of TerraIncognita.}
\label{tab:gen_terra}
\end{table}

\clearpage
\subsection{Personalization}
\begin{table}[htb!]
\centering
\resizebox{\columnwidth}{!}{%
\begin{tabular}{l|c|ccc|c}
\toprule
Method & Target & 0 & 25 & 26 & AVG \\ 
\midrule\midrule
Zero Shot & \multirow{19}{*}{33} & 48.40 $\pm$ 1.47 & 45.19 $\pm$ 2.50 & 48.24 $\pm$ 2.73 & 47.28 $\pm$ 0.73 \\
FedCLIP & & 70.03 $\pm$ 8.40 & 72.44 $\pm$ 11.70 & 69.23 $\pm$ 9.38 & 70.57 $\pm$ 9.77 \\
PromptFL ($M=4$) & & 96.96 $\pm$ 0.74 & 95.83 $\pm$ 1.94 & 96.96 $\pm$ 2.00 & 96.58 $\pm$ 1.49 \\
PromptFL ($M=16$) & & \textbf{97.12 $\pm$ 0.97} & 97.60 $\pm$ 0.48 & \textbf{98.24 $\pm$ 1.47} & \textbf{97.65 $\pm$ 0.72} \\
CoCoOp (All Global) & & 88.46 $\pm$ 7.26 & 89.43 $\pm$ 6.68 & 87.50 $\pm$ 9.36 & 88.46 $\pm$ 7.63 \\
CoCoOp (Meta-Net Random Init) & & 87.18 $\pm$ 6.13 & 87.34 $\pm$ 7.28 & 85.42 $\pm$ 10.59 & 86.64 $\pm$ 8.00 \\
CoCoOp (Server CoOp) & & 94.39 $\pm$ 1.94 & 94.71 $\pm$ 1.44 & 96.32 $\pm$ 1.39 & 95.14 $\pm$ 1.09 \\
\bottomrule
\end{tabular}%
}
\caption{The personalization accuracy of FEMNIST (Target 33, 0).}
\label{tab:per_femnist_1}
\end{table}

\begin{table}[htb!]
\centering
\resizebox{\columnwidth}{!}{%
\begin{tabular}{l|c|ccc|c}
\toprule
Method & Target & 33 & 0 & 26 & AVG \\ 
\midrule\midrule
Zero Shot & \multirow{19}{*}{25} & 50.88 $\pm$ 3.42 & 48.40 $\pm$ 1.47 & 48.24 $\pm$ 2.73 & 49.17 $\pm$ 1.71 \\
FedCLIP & & 67.75 $\pm$ 8.19 & 71.96 $\pm$ 7.93 & 70.35 $\pm$ 10.67 & 70.02 $\pm$ 8.93 \\
PromptFL ($M=4$) & & 95.92 $\pm$ 0.70 & 96.63 $\pm$ 0.49 & 97.28 $\pm$ 0.74 & 96.61 $\pm$ 0.12 \\
PromptFL ($M=16$) & & 96.33 $\pm$ 1.08 & 96.80 $\pm$ 0.56 & 97.92 $\pm$ 0.55 & 97.01 $\pm$ 0.44 \\
CoCoOp (All Global) & & 84.63 $\pm$ 12.12 & 85.58 $\pm$ 10.10 & 85.90 $\pm$ 10.01 & 85.37 $\pm$ 10.72 \\
CoCoOp (Meta-Net Random Init) & & 85.17 $\pm$ 11.83 & 83.33 $\pm$ 9.86 & 84.62 $\pm$ 11.34 & 84.37 $\pm$ 10.95 \\
CoCoOp (Server CoOp) & & 96.33 $\pm$ 1.23 & 94.55 $\pm$ 1.00 & 95.67 $\pm$ 1.66 & 95.52 $\pm$ 0.62 \\
\bottomrule
\end{tabular}%
}
\caption{The personalization accuracy of FEMNIST (Target 25, 26).}
\label{tab:per_femnist_2}
\end{table}

\begin{table}[htb!]
\centering
\resizebox{\columnwidth}{!}{%
\begin{tabular}{l|c|ccc|c}
\toprule
Method & Target & C & P & S & AVG \\ 
\midrule\midrule
Zero Shot & \multirow{19}{*}{A} & 96.94 $\pm$ 0.81 & 99.60 $\pm$ 0.46 & 85.56 $\pm$ 0.99 & 94.03 $\pm$ 0.38 \\
FedCLIP & & 97.01 $\pm$ 1.67 & 99.70 $\pm$ 0.30 & 89.94 $\pm$ 0.46 & 95.55 $\pm$ 0.62 \\
PromptFL ($M=4$) & & 98.36 $\pm$ 0.81 & \textbf{99.80 $\pm$ 0.35} & 92.78 $\pm$ 1.02 & 96.98 $\pm$ 0.56 \\
PromptFL ($M=16$) & & 98.15 $\pm$ 0.50 & 99.70 $\pm$ 0.30 & 92.74 $\pm$ 0.44 & 96.86 $\pm$ 0.23 \\
CoCoOp (All Global) & & 98.08 $\pm$ 0.98 & 99.30 $\pm$ 0.46 & 91.04 $\pm$ 1.91 & 96.14 $\pm$ 1.01 \\
CoCoOp (Meta-Net Random Init) & & 98.15 $\pm$ 1.18 & 99.70 $\pm$ 0.30 & 92.02 $\pm$ 1.54 & 96.62 $\pm$ 0.66 \\
CoCoOp (Server CoOp) & & 97.86 $\pm$ 0.86 & 99.70 $\pm$ 0.30 & 91.42 $\pm$ 0.51 & 96.33 $\pm$ 0.09 \\
\bottomrule
\end{tabular}%
}
\caption{The personalization accuracy of PACS (Target A, C).}
\label{tab:per_pacs_1}
\end{table}

\begin{table}[htb!]
\centering
\resizebox{\columnwidth}{!}{%
\begin{tabular}{l|c|ccc|c}
\toprule
Method & Target & A & C & S & AVG \\ 
\midrule\midrule
Zero Shot & \multirow{19}{*}{P} & 95.35 $\pm$ 0.98 & 96.94 $\pm$ 0.81 & 85.56 $\pm$ 0.99 & 92.62 $\pm$ 0.78 \\
FedCLIP & & 95.11 $\pm$ 1.36 & 98.00 $\pm$ 1.73 & 89.94 $\pm$ 0.26 & 94.35 $\pm$ 0.98 \\
PromptFL ($M=4$) & & 96.82 $\pm$ 0.65 & 98.29 $\pm$ 0.74 & 92.27 $\pm$ 1.28 & 95.80 $\pm$ 0.50 \\
PromptFL ($M=16$) & & 96.41 $\pm$ 1.21 & 98.43 $\pm$ 0.66 & 92.65 $\pm$ 0.32 & 95.83 $\pm$ 0.71 \\
CoCoOp (All Global) & & 96.58 $\pm$ 0.89 & 98.08 $\pm$ 0.77 & 91.04 $\pm$ 1.80 & 95.23 $\pm$ 0.78 \\
CoCoOp (Meta-Net Random Init) & & 96.50 $\pm$ 0.93 & 98.29 $\pm$ 0.98 & 91.85 $\pm$ 1.60 & 95.54 $\pm$ 0.92 \\
CoCoOp (Server CoOp) & & 96.01 $\pm$ 1.77 & 97.86 $\pm$ 0.86 & 91.46 $\pm$ 0.44 & 95.11 $\pm$ 0.76 \\
\bottomrule
\end{tabular}%
}
\caption{The personalization accuracy of PACS (Target P, S).}
\label{tab:per_pacs_2}
\end{table}

\begin{table}[htb!]
\centering
\resizebox{\columnwidth}{!}{%
\begin{tabular}{l|c|ccc|c}
\toprule
Method & Target & C & P & R & AVG \\ 
\midrule\midrule
Zero Shot & \multirow{19}{*}{A} & 65.29 $\pm$ 0.89 & 86.89 $\pm$ 1.61 & 86.83 $\pm$ 0.74 & 79.67 $\pm$ 0.32 \\
FedCLIP & & 69.11 $\pm$ 0.70 & 88.32 $\pm$ 1.43 & 87.60 $\pm$ 0.99 & 81.67 $\pm$ 0.89 \\
PromptFL ($M=4$) & & 74.88 $\pm$ 2.43 & 90.91 $\pm$ 1.09 & 89.40 $\pm$ 0.81 & 85.06 $\pm$ 0.88 \\
PromptFL ($M=16$) & & 74.91 $\pm$ 1.38 & 91.02 $\pm$ 0.72 & 89.17 $\pm$ 0.66 & 85.03 $\pm$ 0.42 \\
CoCoOp (All Global) & & 69.11 $\pm$ 3.52 & 88.95 $\pm$ 0.69 & 88.37 $\pm$ 0.96 & 82.14 $\pm$ 1.05 \\
CoCoOp (Meta-Net Random Init) & & 71.25 $\pm$ 2.25 & 89.10 $\pm$ 1.02 & 88.37 $\pm$ 1.16 & 82.91 $\pm$ 0.58 \\
CoCoOp (Server CoOp) & & 74.88 $\pm$ 2.43 & 90.91 $\pm$ 1.09 & 89.40 $\pm$ 0.81 & 85.06 $\pm$ 0.88 \\
\bottomrule
\end{tabular}%
}
\caption{The personalization accuracy of Office-Home (Target A, C).}
\label{tab:per_oh_1}
\end{table}

\begin{table}[htb!]
\centering
\resizebox{\columnwidth}{!}{%
\begin{tabular}{l|c|ccc|c}
\toprule
Method & Target & A & C & R & AVG \\ 
\midrule\midrule
Zero Shot & \multirow{19}{*}{P} & 77.59 $\pm$ 1.33 & 65.29 $\pm$ 0.89 & 86.83 $\pm$ 0.74 & 76.57 $\pm$ 0.64 \\
FedCLIP & & 78.08 $\pm$ 1.37 & 68.00 $\pm$ 0.82 & 87.49 $\pm$ 0.41 & 77.86 $\pm$ 0.09 \\
PromptFL ($M=4$) & & 82.34 $\pm$ 1.73 & 73.88 $\pm$ 1.69 & 89.44 $\pm$ 0.70 & 81.89 $\pm$ 0.90 \\
PromptFL ($M=16$) & & 83.09 $\pm$ 2.03 & 74.42 $\pm$ 1.98 & 89.82 $\pm$ 0.75 & 82.44 $\pm$ 1.41 \\
CoCoOp (All Global) & & 78.08 $\pm$ 1.45 & 69.15 $\pm$ 2.78 & 88.40 $\pm$ 1.10 & 78.54 $\pm$ 1.52 \\
CoCoOp (Meta-Net Random Init) & & 78.49 $\pm$ 1.34 & 71.02 $\pm$ 2.40 & 88.37 $\pm$ 1.09 & 79.29 $\pm$ 1.32 \\
CoCoOp (Server CoOp) & & 82.34 $\pm$ 1.73 & 73.88 $\pm$ 1.69 & 89.44 $\pm$ 0.70 & 81.89 $\pm$ 0.90 \\
\bottomrule
\end{tabular}%
}
\caption{The personalization accuracy of Office-Home (Target P, R).}
\label{tab:per_oh_2}
\end{table}

\begin{table}[htb!]
\centering
\resizebox{\columnwidth}{!}{%
\begin{tabular}{l|c|ccc|c}
\toprule
Method & Target & L & S & V & AVG \\ 
\midrule\midrule
Zero Shot & \multirow{19}{*}{C} & 65.22 $\pm$ 2.72 & 70.96 $\pm$ 0.33 & 84.79 $\pm$ 1.07 & 74.06 $\pm$ 1.24 \\
FedCLIP & & 70.37 $\pm$ 2.68 & 81.35 $\pm$ 1.64 & 86.96 $\pm$ 1.31 & 79.56 $\pm$ 1.33 \\
PromptFL ($M=4$) & & 72.44 $\pm$ 0.71 & 84.75 $\pm$ 2.97 & 88.99 $\pm$ 1.44 & 82.06 $\pm$ 1.27 \\
PromptFL ($M=16$) & & 73.07 $\pm$ 0.98 & 83.69 $\pm$ 1.58 & 88.39 $\pm$ 1.38 & 81.72 $\pm$ 0.66 \\
CoCoOp (All Global) & & 72.69 $\pm$ 2.29 & 82.37 $\pm$ 1.96 & 87.90 $\pm$ 3.33 & 80.99 $\pm$ 0.96 \\
CoCoOp (Meta-Net Random Init) & & 78.66 $\pm$ 1.44 & 84.70 $\pm$ 1.01 & 88.10 $\pm$ 1.13 & 83.82 $\pm$ 1.13 \\
CoCoOp (Server CoOp) & & 79.35 $\pm$ 0.66 & 83.23 $\pm$ 0.93 & 88.39 $\pm$ 1.57 & 83.66 $\pm$ 0.58 \\
\bottomrule
\end{tabular}%
}
\caption{The personalization accuracy of VLCS (Target C, L).}
\label{tab:per_vlcs_1}
\end{table}

\begin{table}[htb!]
\centering
\resizebox{\columnwidth}{!}{%
\begin{tabular}{l|c|ccc|c}
\toprule
Method & Target & C & L & V & AVG \\ 
\midrule\midrule
Zero Shot & \multirow{19}{*}{S} & \textbf{100.00 $\pm$ 0.00} & 65.22 $\pm$ 2.72 & 84.79 $\pm$ 1.07 & 83.34 $\pm$ 1.11 \\
FedCLIP & & 100.00 $\pm$ 0.00 & 74.45 $\pm$ 1.83 & 87.06 $\pm$ 0.73 & 87.17 $\pm$ 0.72 \\
PromptFL ($M=4$) & & 99.65 $\pm$ 0.61 & 76.46 $\pm$ 0.98 & 88.74 $\pm$ 1.17 & 88.28 $\pm$ 0.37 \\
PromptFL ($M=16$) & & 99.77 $\pm$ 0.20 & 76.96 $\pm$ 1.28 & 88.00 $\pm$ 0.74 & 88.24 $\pm$ 0.56 \\
CoCoOp (All Global) & & \textbf{100.00 $\pm$ 0.00} & 77.09 $\pm$ 0.47 & 87.06 $\pm$ 2.64 & 88.05 $\pm$ 0.73 \\
CoCoOp (Meta-Net Random Init) & & \textbf{100.00 $\pm$ 0.00} & 78.21 $\pm$ 1.23 & 88.05 $\pm$ 1.30 & 88.75 $\pm$ 0.78 \\
CoCoOp (Server CoOp) & & 99.77 $\pm$ 0.20 & 79.53 $\pm$ 0.85 & 88.20 $\pm$ 1.78 & 89.17 $\pm$ 0.87 \\
\bottomrule
\end{tabular}%
}
\caption{The personalization accuracy of VLCS (Target S, V).}
\label{tab:per_vlcs_2}
\end{table}

\begin{table}[htb!]
\centering
\resizebox{\columnwidth}{!}{%
\begin{tabular}{l|c|ccc|c}
\toprule
Method & Target & L43 & L46 & L100 & AVG \\ 
\midrule\midrule
Zero Shot & \multirow{19}{*}{L38} & 29.90 $\pm$ 1.95 & 19.59 $\pm$ 1.31 & 13.47 $\pm$ 0.52 & 20.99 $\pm$ 0.43 \\
FedCLIP & & 44.17 $\pm$ 9.98 & 50.91 $\pm$ 7.83 & 62.60 $\pm$ 19.22 & 52.56 $\pm$ 12.32 \\
PromptFL ($M=4$) & & 54.71 $\pm$ 1.85 & 58.16 $\pm$ 0.37 & 67.97 $\pm$ 2.87 & 60.28 $\pm$ 1.66 \\
PromptFL ($M=16$) & & 53.09 $\pm$ 1.39 & 58.81 $\pm$ 0.28 & 67.69 $\pm$ 2.54 & 59.87 $\pm$ 0.56 \\
CoCoOp (All Global) & & 47.29 $\pm$ 0.66 & 54.02 $\pm$ 0.32 & 65.34 $\pm$ 5.52 & 55.55 $\pm$ 1.75 \\
CoCoOp (Meta-Net Random Init) & & 58.48 $\pm$ 4.88 & 59.36 $\pm$ 0.53 & 78.63 $\pm$ 1.40 & 65.49 $\pm$ 1.52 \\
CoCoOp (Server CoOp) & & 59.31 $\pm$ 0.77 & 58.00 $\pm$ 5.81 & 76.07 $\pm$ 3.18 & 64.46 $\pm$ 1.69 \\
\bottomrule
\end{tabular}%
}
\caption{The personalization accuracy of TerraIncognita (Target L38, L43).}
\label{tab:per_terra_1}
\end{table}

\begin{table}[htb!]
\centering
\resizebox{\columnwidth}{!}{%
\begin{tabular}{l|c|ccc|c}
\toprule
Method & Target & L38 & L43 & L100 & AVG \\ 
\midrule\midrule
Zero Shot & \multirow{19}{*}{L46} & 13.35 $\pm$ 0.76 & 29.90 $\pm$ 1.95 & 13.47 $\pm$ 0.52 & 18.91 $\pm$ 0.68 \\
FedCLIP & & 61.53 $\pm$ 13.73 & 37.16 $\pm$ 6.02 & 66.77 $\pm$ 18.25 & 55.15 $\pm$ 12.60 \\
PromptFL ($M=4$) & & 64.43 $\pm$ 1.79 & 48.98 $\pm$ 2.73 & 69.44 $\pm$ 2.96 & 60.95 $\pm$ 0.59 \\
PromptFL ($M=16$) & & 63.22 $\pm$ 2.08 & 48.65 $\pm$ 2.09 & 69.92 $\pm$ 3.23 & 60.60 $\pm$ 0.21 \\
CoCoOp (All Global) & & 64.67 $\pm$ 1.51 & 38.91 $\pm$ 5.01 & 67.49 $\pm$ 2.42 & 57.02 $\pm$ 1.30 \\
CoCoOp (Meta-Net Random Init) & & 74.14 $\pm$ 0.32 & 58.56 $\pm$ 4.42 & 78.33 $\pm$ 1.44 & 70.34 $\pm$ 1.21 \\
CoCoOp (Server CoOp) & & 73.29 $\pm$ 2.06 & 59.14 $\pm$ 1.30 & 77.20 $\pm$ 3.10 & 69.88 $\pm$ 1.50 \\
\bottomrule
\end{tabular}%
}
\caption{The personalization accuracy of TerraIncognita (Target L46, L100).}
\label{tab:per_terra_2}
\end{table}

\clearpage
\subsection{Comprehensive}
\begin{table}[htb!]
\centering
\resizebox{\columnwidth}{!}{%
\begin{tabular}{l|cccc|c}
\toprule
Method    & 33    & 0     & 25    & 26    & AVG   \\ \midrule\midrule
Zero Shot & $50.88 \pm 3.42$ & $48.40 \pm 1.47$ & $45.19 \pm 2.50$ & $48.24 \pm 2.73$ & $48.18 \pm 1.31$ \\
FedCLIP   & 70.07 $\pm$ 9.35 & 70.58 $\pm$ 8.74 & 70.55 $\pm$ 9.69 & 70.82 $\pm$ 9.57 & 70.51 $\pm$ 9.30 \\
PromptFL ($M=4$) & 96.01 $\pm$ 1.56  & 96.13 $\pm$ 0.49 & 96.66 $\pm$ 0.30 & 96.82 $\pm$ 0.67 & 96.40 $\pm$ 0.74 \\ 
PromptFL ($M=16$) & \textbf{97.12 $\pm$ 0.64} & 96.87 $\pm$ 0.53 & 96.92 $\pm$ 0.52 & 96.95 $\pm$ 0.27 & 96.96 $\pm$ 0.48 \\ 
CoCoOp (All Global) & 87.91 $\pm$ 8.43 & 87.29 $\pm$ 8.86 & 85.90 $\pm$ 9.89 & 87.94 $\pm$ 7.46 & 87.26 $\pm$ 8.65 \\
CoCoOp (Meta-Net Random Init) & 70.29 $\pm$ 8.78 & 69.58 $\pm$ 10.48 & 69.17 $\pm$ 10.76 & 70.31 $\pm$ 9.64 & 69.84 $\pm$ 9.87 \\
CoCoOp (Server CoOp) & 75.13 $\pm$ 1.06 & 76.02 $\pm$ 0.82 & 76.04 $\pm$ 2.84 & 73.86 $\pm$ 1.03 & 75.26 $\pm$ 0.82 \\
\bottomrule
\end{tabular}%
}
\caption{The comprehensive accuracy of FEMNIST.}
\label{tab:comp_femnist}
\end{table}

\begin{table}[htb!]
\centering
\resizebox{\columnwidth}{!}{%
\begin{tabular}{l|cccc|c}
\toprule
Method    & A     & C     & P     & S     & AVG   \\ \midrule\midrule
Zero Shot & $95.35 \pm 0.98$ & \textbf{96.94 $\pm$ 0.81} & \textbf{99.60 $\pm$ 0.46} & $85.56 \pm 0.99$ & $94.36 \pm 0.48$ \\
FedCLIP   & 95.50 $\pm$ 0.75 & 95.61 $\pm$ 0.62 & 95.59 $\pm$ 0.62 & 94.99 $\pm$ 1.08 & 95.42 $\pm$ 0.76 \\
PromptFL ($M=4$) & 96.56 $\pm$ 0.42 & 96.74 $\pm$ 0.30 & 96.72 $\pm$ 0.34 & 94.92 $\pm$ 0.44 & 96.23 $\pm$ 0.37 \\ 
PromptFL ($M=16$) & 96.16 $\pm$ 0.36 & 96.52 $\pm$ 0.22 & 96.73 $\pm$ 0.54 & 95.31 $\pm$ 0.31 & 96.18 $\pm$ 0.33 \\ 
CoCoOp (All Global) & 95.92 $\pm$ 1.05 & 96.11 $\pm$ 0.81 & 96.15 $\pm$ 0.74 & 95.06 $\pm$ 0.65 & 95.81 $\pm$ 0.80 \\
CoCoOp (Meta-Net Random Init) & 87.99 $\pm$ 3.92 & 91.30 $\pm$ 2.33 & 87.35 $\pm$ 4.49 & 91.64 $\pm$ 2.48 & 89.57 $\pm$ 3.27 \\
CoCoOp (Server CoOp) & 93.68 $\pm$ 0.22 & 90.47 $\pm$ 6.63 & 93.69 $\pm$ 1.32 & 93.60 $\pm$ 3.39 & 92.86 $\pm$ 2.22 \\
\bottomrule
\end{tabular}%
}
\caption{The comprehensive accuracy of PACS.}
\label{tab:comp_pacs}
\end{table}

\begin{table}[htb!]
\centering
\resizebox{\columnwidth}{!}{%
\begin{tabular}{l|cccc|c}
\toprule
Method    & A     & C     & P     & R     & AVG   \\ \midrule\midrule
Zero Shot & 77.59 $\pm$ 1.33 & 65.29 $\pm$ 0.89 & \textbf{86.89 $\pm$ 1.61} & \textbf{86.83 $\pm$ 0.74} & 79.15 $\pm$ 0.40 \\
FedCLIP   & 80.59 $\pm$ 0.41 & 80.06 $\pm$ 0.09 & 79.88 $\pm$ 0.42 & 79.94 $\pm$ 0.49 & 80.12 $\pm$ 0.17 \\
PromptFL ($M=4$) & 83.14 $\pm$ 1.12 & 82.66 $\pm$ 0.69 & 83.38 $\pm$ 0.62 & 84.04 $\pm$ 1.02 & 83.31 $\pm$ 0.86 \\
PromptFL ($M=16$) & 83.74 $\pm$ 0.42 & 82.71 $\pm$ 0.57 & 83.95 $\pm$ 0.95 & 84.06 $\pm$ 0.96 & 83.61 $\pm$ 0.69 \\
CoCoOp (All Global) & 81.04 $\pm$ 1.26 & 80.21 $\pm$ 1.01 & 81.00 $\pm$ 0.94 & 81.36 $\pm$ 1.15 & 80.90 $\pm$ 1.07 \\
CoCoOp (Meta-Net Random Init) & 77.42 $\pm$ 1.92 & 76.59 $\pm$ 0.92 & 76.35 $\pm$ 1.70 & 77.00 $\pm$ 1.05 & 76.84 $\pm$ 1.30 \\
CoCoOp (Server CoOp) & 83.14 $\pm$ 1.12 & 82.66 $\pm$ 0.69 & 83.38 $\pm$ 0.62 & 84.04 $\pm$ 1.02 & 83.31 $\pm$ 0.86 \\
\bottomrule
\end{tabular}%
}
\caption{The comprehensive accuracy of Office-Home.}
\label{tab:comp_oh}
\end{table}

\begin{table}[htb!]
\centering
\resizebox{\columnwidth}{!}{%
\begin{tabular}{l|cccc|c}
\toprule
Method    & C     & L     & S     & V     & AVG   \\ \midrule\midrule
Zero Shot & \textbf{100.00 $\pm$ 0.00} & 65.22 $\pm$ 2.72 & 70.96 $\pm$ 0.33 & 84.79 $\pm$ 1.07 & 80.59 $\pm$ 0.96 \\
FedCLIP   & 84.49 $\pm$ 1.04 & 82.93 $\pm$ 0.54 & 83.75 $\pm$ 0.59 & 84.03 $\pm$ 0.40 & 83.80 $\pm$ 0.63 \\
PromptFL ($M=4$) & 86.31 $\pm$ 0.88 & 84.81 $\pm$ 0.30 & 85.95 $\pm$ 0.51 & 85.99 $\pm$ 0.72 & 85.77 $\pm$ 0.57 \\
PromptFL ($M=16$) & 86.08 $\pm$ 0.39 & 83.82 $\pm$ 0.18 & 85.62 $\pm$ 1.05 & 85.55 $\pm$ 0.90 & 85.27 $\pm$ 0.54 \\
CoCoOp (All Global) & 85.71 $\pm$ 0.67 & 83.60 $\pm$ 0.93 & 84.62 $\pm$ 1.47 & 85.11 $\pm$ 1.85 & 84.76 $\pm$ 1.01 \\
CoCoOp (Meta-Net Random Init) & 83.82 $\pm$ 1.13 & \textbf{90.97 $\pm$ 0.85} & \textbf{88.75 $\pm$ 0.78} & 87.76 $\pm$ 0.98 & 79.91 $\pm$ 1.06 \\
CoCoOp (Server CoOp) & 85.06 $\pm$ 4.23 & 81.05 $\pm$ 2.72 & 81.97 $\pm$ 0.78 & 83.99 $\pm$ 0.49 & 83.02 $\pm$ 1.46 \\
\bottomrule
\end{tabular}%
}
\caption{The comprehensive accuracy of VLCS.}
\label{tab:comp_vlcs}
\end{table}

\begin{table}[htb!]
\centering
\resizebox{\columnwidth}{!}{%
\begin{tabular}{l|cccc|c}
\toprule
Method    & L38   & L43   & L46   & L100  & AVG   \\ \midrule\midrule
Zero Shot & 13.35 $\pm$ 0.76 & 29.90 $\pm$ 1.95 & 19.59 $\pm$ 1.31 & 13.47 $\pm$ 0.52 & 19.08 $\pm$ 0.26 \\
FedCLIP   & 47.74 $\pm$ 8.38 & 49.10 $\pm$ 9.72 & 49.03 $\pm$ 9.81 & 50.89 $\pm$ 8.74 & 49.19 $\pm$ 9.11 \\
PromptFL ($M=4$) & 54.84 $\pm$ 0.50 & 54.42 $\pm$ 1.80 & 54.96 $\pm$ 0.91 & 53.49 $\pm$ 1.57 & 54.43 $\pm$ 0.75 \\
PromptFL ($M=16$) & 54.23 $\pm$ 2.03 & 55.16 $\pm$ 1.98 & 54.27 $\pm$ 0.98 & 52.52 $\pm$ 0.58 & 54.04 $\pm$ 1.07 \\
CoCoOp (All Global) & 51.24 $\pm$ 3.28 & 51.93 $\pm$ 0.56 & 52.48 $\pm$ 0.73 & 50.32 $\pm$ 1.83 & 51.49 $\pm$ 0.16 \\
CoCoOp (Meta-Net Random Init) & 60.28 $\pm$ 1.33 & 59.75 $\pm$ 1.47 & 59.14 $\pm$ 1.49 & 54.14 $\pm$ 1.61 & 58.33 $\pm$ 0.86 \\
CoCoOp (Server CoOp) & 56.88 $\pm$ 5.74 & 58.61 $\pm$ 1.00 & 58.45 $\pm$ 0.79 & 52.65 $\pm$ 1.74 & 56.65 $\pm$ 2.22 \\
\bottomrule
\end{tabular}%
}
\caption{The comprehensive accuracy of TerraIncognita.}
\label{tab:comp_terra}
\end{table}

\clearpage
\twocolumn

{
    \small
    \bibliographystyle{ieeenat_fullname}
    \bibliography{main}
}